\documentclass[runningheads]{llncs}

\usepackage{eccv}

\usepackage{multirow, makecell}

\usepackage{eccvabbrv}

\usepackage{graphicx}
\usepackage{booktabs}

\usepackage[accsupp]{axessibility}  

\usepackage[pagebackref,breaklinks,colorlinks,citecolor=eccvblue]{hyperref}

\usepackage{orcidlink}

\usepackage{microtype} 

\begin{document}

\title{ViewSplat: View-Adaptive 3D Gaussian Splatting for Feed-Forward Synthesis}

\titlerunning{View-Adaptive Gaussian Splatting}

\author{Moonyeon Jeong\inst{1} \and Seunggi Min\inst{1} \and Suhyeon Lee\inst{2}\textsuperscript{,$\dagger$} \and Hongje Seong\inst{1}\textsuperscript{,$\dagger$}}

\authorrunning{M.~Jeong et al.}

\institute{\textsuperscript{1} University of Seoul\quad \textsuperscript{2} Korea Electronics Technology Institute
\email{\{jmy06051,minsk0725,hjseong\}@uos.ac.kr\quad suhyeon@keti.re.kr}\\
\url{https://cvlab-uos.github.io/ViewSplat}
}

\maketitle

\begin{abstract}
We present ViewSplat, a view-adaptive 3D Gaussian splatting network for novel view synthesis from unposed images.
While recent feed-forward 3D Gaussian splatting has significantly accelerated 3D scene reconstruction by bypassing per-scene optimization, a fundamental fidelity gap remains.
We attribute this gap to the limited capacity of single-step feed-forward networks to regress static Gaussian primitives that satisfy all viewpoints.
To address this limitation, we shift the paradigm from static primitive regression to view-adaptive splatting.
Instead of a rigid Gaussian representation, our pipeline learns a view-adaptive latent representation.
Specifically, ViewSplat initially predicts base Gaussian primitives alongside the weights of scene-conditioned View MLPs.
During rendering, these MLPs take target-view coordinates as input and predict view-dependent residual updates for each Gaussian attribute (\ie, 3D position, scale, rotation, opacity, and color).
This mechanism, which we term view-adaptive splatting, allows each primitive to rectify initial estimation errors, effectively capturing high-fidelity appearances.
Extensive experiments demonstrate that ViewSplat achieves state-of-the-art fidelity while maintaining fast inference and real-time rendering; our large backbone variant runs at 15~FPS during inference and 90~FPS during rendering.
\end{abstract}

\let\thefootnote\relax\footnotetext{\textsuperscript{$\dagger$}Co-corresponding authors.}

\section{Introduction}
\label{sec:intro}
3D Gaussian Splatting (3DGS)~\cite{kerbl20233dgs} has recently emerged as a powerful representation in 3D computer vision, enabling high-fidelity real-time rendering.
While traditional 3DGS frameworks rely on per-scene optimization, recent efforts have focused on generalizable feed-forward architectures \cite{charatan2024pixelsplat,chen2024mvsplat,xu2024grm,tang2024lgm,zhang2024gs-lrm}.
These models predict scene representations directly from a few input images, enabling instant 3D reconstruction of unseen scenes and high-quality novel view synthesis (NVS) without costly target-scene optimization.

\begin{figure}[t]
    \centering
    \includegraphics[width=1\linewidth]{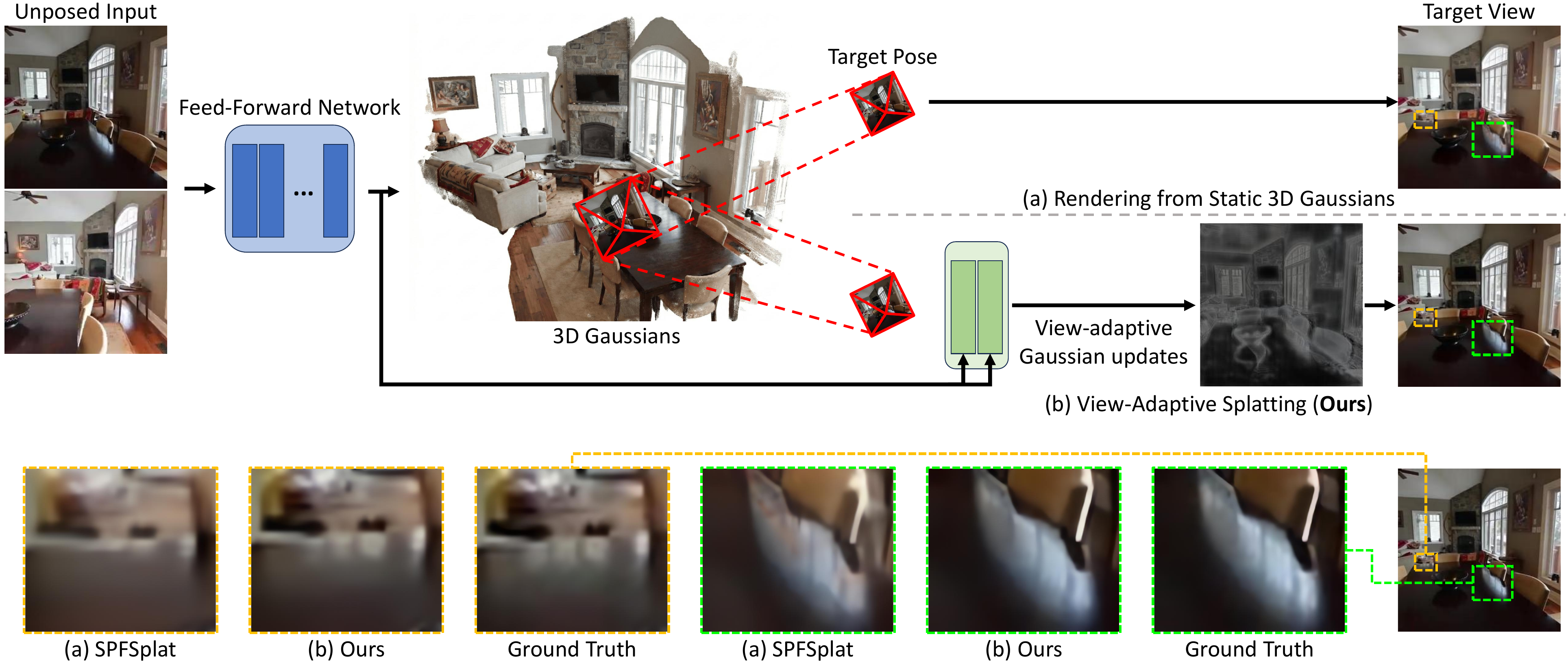}
    \caption{Given unposed images as input, the feed-forward network reconstructs 3D Gaussians. While (a) rendering from static 3D Gaussians often suffers from blurred details, (b) our method introduces view-adaptive Gaussian updates conditioned on the target-view pose to refine the Gaussians on the fly. As shown in the bottom panels, our approach improves the reconstruction of fine-grained details (\eg, sharp edges and specular reflections) compared to SPFSplat~\cite{huang2025spfsplat}.
    }
    \label{fig:intro}
\end{figure}

A major bottleneck in these generalizable pipelines is the heavy reliance on accurate camera poses, typically precomputed via Structure-from-Motion (SfM)~\cite{schonberger2016sfmrevisited}.
To address this, recent works~\cite{smart2024splatt3r,kang2025selfsplat,huang2025spfsplat,huang2025spfsplatv2} leverage 3D foundation models---such as DUSt3R~\cite{wang2024dust3r}, MASt3R~\cite{leroy2024mast3r}, and VGGT~\cite{wang2025vggt}---to facilitate pose-free reconstruction.
These frameworks enable the joint estimation of camera parameters and 3D Gaussian primitives~\cite{jiang2025anysplat,huang2025spfsplat}, delivering robust 3D reconstructions directly from uncalibrated, in-the-wild images.

Despite these advancements, a significant fidelity gap remains between feed-forward approaches and optimization-based methods~\cite{jiang2024gaussianshader,zhou2025rtr-gs,yang2024spec-gaussian,liu2024mirrorgaussian,lu2024scaffold-gs}. We attribute this gap to the limited capacity of feed-forward networks to regress static Gaussian primitives that satisfy all potential viewpoints. These models typically employ a single-step regression to predict all Gaussian attributes (\ie, 3D position, scale, rotation, opacity, and spherical harmonics (SH) coefficients). However, regressing optimal representations is an ill-posed task for a single forward pass, especially when dealing with complex geometries or non-Lambertian surfaces. While optimization-based methods can refine these attributes over thousands of iterations to capture sharp specularities and view-dependent SH coefficients, feed-forward models lack the capacity to \textit{rectify} their initial predictions during inference.

To address this limitation, we propose ViewSplat, a novel framework that shifts the paradigm from static primitive regression to view-adaptive splatting (\cref{fig:intro}).
The key insight behind ViewSplat is that achieving high-fidelity 2D rendering from a \textit{specific} view is relatively straightforward, provided the underlying 3D primitives can be tailored to that viewpoint. We leverage this observation by shifting the network’s objective: instead of struggling to predict universal 3D Gaussians, the network focuses on generating view-specific updates that are far more feasible to learn.
This shift allows our model to rectify initial errors and capture complex view-dependent details that a static model would otherwise miss. Building upon a pose-free feed-forward architecture~\cite{huang2025spfsplat,huang2025spfsplatv2}, our model introduces a mechanism that updates Gaussian primitives for every target viewpoint on-the-fly.

Specifically, ViewSplat predicts two components for a given scene: a set of base Gaussian primitives and the weights of scene-conditioned View MLPs. Unlike standard MLPs with fixed parameters, these View MLPs are conditioned on the input scene context. During the rendering stage, these MLPs take the target-view coordinates as input and predict view-dependent residual updates for all Gaussian attributes, including 3D position, scale, rotation, opacity, and SH coefficients. This allows each primitive to undergo geometric and photometric refinement tailored to the specific viewpoint. Consequently, ViewSplat can effectively compensate for initial estimation errors and synthesize high-frequency, non-Lambertian effects that are typically lost in static feed-forward representations.
Our approach maintains the efficiency required for practical applications, achieving fast inference (15 FPS) and high-speed rendering (90 FPS) for the large-backbone variant while significantly outperforming existing generalizable models.
Our main contributions are summarized as follows:
\begin{itemize}
    \item We identify the static primitive bottleneck in feed-forward 3DGS and propose view-adaptive representations for high-fidelity synthesis.
    \item We introduce ViewSplat, which utilizes scene-conditioned View MLPs to predict view-dependent residuals for all Gaussian attributes, enabling the model to rectify geometric and appearance errors on-the-fly.
    \item We demonstrate that ViewSplat achieves state-of-the-art performance on standard benchmarks, effectively capturing complex view-dependent effects without the need for costly per-scene optimization or precomputed poses.
\end{itemize}

\section{Related Work}
\subsection{Feed-Forward 3D Gaussian Splatting}
The landscape of 3DGS~\cite{kerbl20233dgs} has recently diversified, with significant research efforts increasingly directed toward generalizable feed-forward architectures that enable instant 3D reconstruction and NVS.
Early generalizable 3DGS methods are pose-required, relying on accurate camera parameters to aggregate multi-view features or construct cost volumes. For instance, pixelSplat~\cite{charatan2024pixelsplat} and MVSplat~\cite{chen2024mvsplat} leverage epipolar transformers for depth estimation and Gaussian primitive prediction. Similarly, large-scale models such as LGM~\cite{tang2024lgm}, GRM~\cite{xu2024grm}, and GS-LRM~\cite{zhang2024gs-lrm} encode camera poses via Pl\"ucker ray embeddings. While effective, such methods rely on precomputed poses through SfM~\cite{schonberger2016sfmrevisited}, which is computationally expensive and often unreliable in sparse-view scenarios.

To overcome these constraints, pose-free feed-forward methods have emerged. Supervised pose-free models like Splatt3R~\cite{smart2024splatt3r} and NoPoSplat~\cite{ye2025noposplat} predict 3D Gaussians in a canonical space without inference-time poses, yet they still require ground-truth poses for training supervision. More recently, self-supervised approaches such as SelfSplat~\cite{kang2025selfsplat}, PF3plat \cite{hong2025pf3plat}, SPFSplat \cite{huang2025spfsplat}, and SPFSplatV2~\cite{huang2025spfsplatv2} have effectively eliminated this dependency. In particular, SPFSplat and SPFSplatV2 leverage large-scale 3D foundation model backbones, such as MASt3R~\cite{leroy2024mast3r} and VGGT~\cite{wang2025vggt} to jointly estimate camera poses and 3D geometry in a unified end-to-end pipeline.


Despite these advancements, existing feed-forward 3DGS models are fundamentally constrained by static primitive regression. By predicting a fixed set of Gaussian attributes in a single pass, these methods struggle to reconcile conflicting visual information across diverse viewpoints, leading to blurred details and a failure to capture non-Lambertian effects like sharp specularities. Our ViewSplat breaks this bottleneck by shifting from rigid regression to view-adaptive refinement. Built upon the self-supervised, pose-free feed-forward frameworks~\cite{huang2025spfsplat,huang2025spfsplatv2}, we introduce a view-dependent update mechanism that \textit{rectifies} Gaussian parameters on the fly for each target viewpoint, enabling high-fidelity synthesis that was previously only attainable through per-scene optimization.

\subsection{View-Dependent 3D Representation}
Modeling view-dependent effects is a long-standing challenge in neural rendering. In Neural Radiance Fields (NeRF)~\cite{mildenhall2020nerf}, the standard approach involves conditioning the radiance MLP on the viewing direction to capture view-dependent color. However, this often struggles to represent complex surface properties, particularly non-Lambertian effects and sharp specularities. To address this limitation, several works~\cite{boss2021nerd,boss2021neural-pil,srinivasan2021nerv,Zhang2021nerfactor} decompose visual appearance into intrinsic properties---such as lighting and materials---typically parameterized via Bidirectional Reflectance Distribution Functions (BRDFs). Ref-NeRF~\cite{verbin2022ref-nerf} further enhances specular fidelity by utilizing reflection vectors. While effective, these implicit methods often require auxiliary geometric priors, such as those derived from signed distance functions (SDFs)~\cite{liu2023nero,liang2023envidr}, to obtain high-quality normals for accurate shading.

The introduction of 3DGS~\cite{kerbl20233dgs} has opened new avenues for explicit view-dependent modeling. While 3DGS originally employs low-order SH to represent view-dependent color, SH coefficients frequently lack the frequency bandwidth necessary to capture sharp specularities. To overcome this limitation, recent extensions like GaussianShader~\cite{jiang2024gaussianshader} and RTR-GS~\cite{zhou2025rtr-gs} integrate explicit shading models into the Gaussian framework. Other approaches, such as Spec-Gaussian~\cite{yang2024spec-gaussian}, introduce anisotropic spherical Gaussians to provide more flexible parameterization than traditional SH. MirrorGaussian~\cite{liu2024mirrorgaussian} tackles the specific challenge of mirror reflections by incorporating symmetry-based Gaussian reflections. Beyond appearance, Scaffold-GS~\cite{lu2024scaffold-gs} explores structural stabilization of Gaussian attributes across viewpoints using an anchor-based representation.

While these advancements have significantly improved the fidelity of view-dependent rendering, they remain predominantly tied to per-scene optimization. This dependency results in high computational costs and a lack of generalization to unseen scenes. In contrast, ViewSplat addresses view dependence within a feed-forward framework by introducing viewpoint-conditioned attribute modulation. 
This allows for efficient, high-fidelity reconstruction and synthesis of complex view-dependent effects without the need for scene-specific training or costly iterative optimization.

\begin{figure}[t]
    \centering
    \includegraphics[width=1\linewidth]{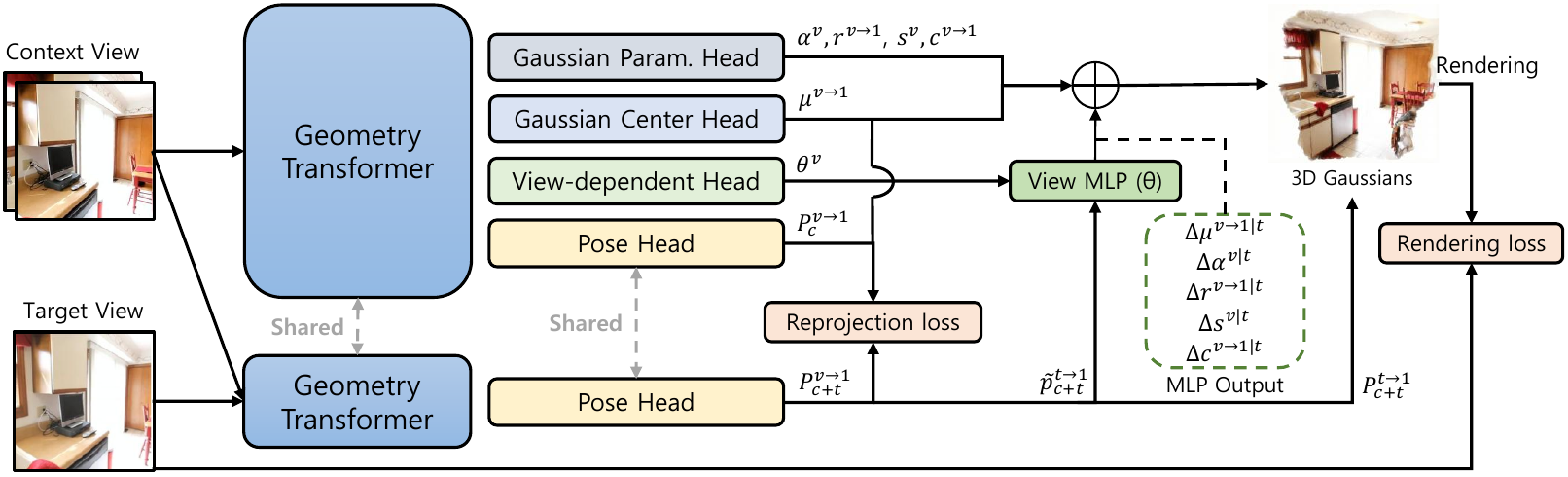}
    \caption{Architecture of ViewSplat. Built upon a shared geometry transformer backbone \cite{leroy2024mast3r,wang2025vggt}, our framework simultaneously predicts canonical 3D Gaussians and camera poses from unposed images using Gaussian heads and a pose head. To accurately capture view-dependent effects, a view-dependent head generates per-pixel View MLPs, which take the target-view pose as input to predict residual offsets for Gaussians. These offsets are then applied to refine the canonical Gaussians during rendering.}
    \label{fig:architecture}
\end{figure}

\section{Method}
We propose ViewSplat, a view-dependent feed-forward 3D Gaussian splatting framework that jointly reconstructs 3D Gaussians and estimates camera poses directly from multi-view images without requiring ground-truth poses. Given a set of images, the network predicts the 3D Gaussian parameters in a canonical space defined by the first input image, which serves as the global reference coordinate frame. Simultaneously, it estimates the camera pose for each view, representing the canonical-to-camera transformation. Leveraging this relative pose information, our model predicts view-dependent residual offsets for the Gaussian parameters. These offsets are added to the canonical Gaussians to refine their attributes, enabling rendering of view-dependent effects such as specular highlights.

An overview of our architecture is illustrated in \cref{fig:architecture}. We first present the preliminaries of our pose-free Gaussian reconstruction framework based on SPFSplat~\cite{huang2025spfsplat,huang2025spfsplatv2} in \cref{sec:preliminaries}. We then introduce our proposed view-dependent head in \cref{sec:vdhead} and detail the view-dependent Gaussian refinement process in \cref{sec:vdrefine}. Finally, we describe the training objective, including the rendering and reprojection losses, in \cref{sec:loss}.

\subsection{Preliminaries}
\label{sec:preliminaries}
Our framework builds upon prior pose-free Gaussian reconstruction architectures, including SPFSplat and its subsequent variants~\cite{huang2025spfsplat,huang2025spfsplatv2}. These models adopt geometry transformer backbones, such as MASt3R~\cite{leroy2024mast3r} and VGGT~\cite{wang2025vggt}, along with pose and Gaussian prediction heads to reconstruct a 3D scene from $N$ unposed images $\{I^v\}_{v=1}^N$.

\subsubsection{Geometry Transformer.}
A ViT-based encoder-decoder architecture~\cite{dosovitskiy2021vit} is employed to extract multi-view feature tokens. The ViT encoder processes patchified RGB inputs through self-attention, after which the decoder enables cross-view interaction to aggregate geometric and appearance information across the $N$ encoded views.

The specific interaction mechanism is determined by the backbone. MASt3R-based models employ pairwise cross-attention between views, whereas the VGGT-based V2-L variant performs global self-attention over a unified token sequence. In SPFSplatV2 and V2-L, a masked attention strategy is additionally introduced in the decoder to improve the efficiency of camera pose estimation and geometry reconstruction. The resulting tokens, enriched with global 3D context, serve as the foundation for both the baseline Gaussian reconstruction and our proposed view-dependent refinement.

\subsubsection{Canonical 3D Gaussian Reconstruction.}
To reconstruct the 3D scene, two Dense Prediction Transformer (DPT)-based heads \cite{ranftl2021dpt} predict 3D Gaussian parameters in a dense, per-pixel manner. The first DPT head, referred to as the Gaussian center head, processes the decoded tokens to predict per-pixel 3D coordinates that define the Gaussian centers. The second head, referred to as the Gaussian parameter head, estimates the remaining Gaussian attributes for each pixel. These 3D Gaussians are aligned in a canonical space, where the first input view $I_1$ serves as the global reference coordinate frame. Specifically, the predicted Gaussians are parameterized as:
\begin{equation}
    \{\mathcal{G}^{v \to 1}\}_{v=1, \dots, N} = \{ (\mu_j^{v \to 1}, \alpha_j^v, r_j^{v \to 1}, s_j^v, c_j^{v \to 1}) \}_{j=1, \dots, H \times W},
    \label{eq:gaussianparams}
\end{equation}
where each Gaussian primitive consists of a center position $\mu \in \mathbb{R}^3$, a rotation quaternion $r \in \mathbb{R}^4$, a scale vector $s \in \mathbb{R}^3$, an opacity value $\alpha \in \mathbb{R}$, and SH coefficients $c$. Following DPT, high-resolution skip connections are incorporated to preserve fine-grained spatial details from the input images.

\subsubsection{Pose Estimation.}
For pose-free rendering, the pose head estimates the camera poses in a single feed-forward step. By taking the concatenated and pooled representations from the encoder and decoder, a lightweight MLP outputs a 10-dimensional relative pose comprising a 6D rotation representation and a 4D homogeneous translation vector $[t_x, t_y, t_z, 1]^\top$, following \cite{chen2024marepo}. This pose is then converted into a homogeneous transformation matrix $P^{v \to 1} \in \mathbb{R}^{4 \times 4}$, which is defined as:
\begin{equation}
    P^{v \to 1} = \begin{bmatrix} R^{v \to 1} & t^{v \to 1} \\ 
    \mathbf{0}^\top & 1 \end{bmatrix},
    \label{eq:contextpose}
\end{equation}
where $R^{v \to 1} \in \mathbb{R}^{3 \times 3}$ is the rotation matrix and $t^{v \to 1} \in \mathbb{R}^3$ is the translation vector. This matrix is the relative transformation of any input view $I_v$ to the canonical reference view $I_1$. To establish a consistent global coordinate system, the first input view $I_1$ is defined as the origin, explicitly assigning its canonical pose as:
\begin{equation}
    P^{1 \to 1} = \begin{bmatrix} I & \mathbf{0} \\ \mathbf{0}^\top & 1
    \end{bmatrix},
    \label{eq:canonicalpose}
\end{equation}
where $I$ represents the identity matrix and $\mathbf{0}$ denotes the zero translation vector.

During training, to synthesize novel views without ground-truth poses, the model additionally predicts the poses of target views, denoted as $P^{t \to 1}$. As illustrated in \cref{fig:architecture}, this target pose estimation leverages features from both context and target views to better capture the global scene structure. Importantly, to prevent information leakage, the initial Gaussian reconstruction is strictly conditioned on the context views, and remains independent of the target pose prediction.

\subsection{View-Dependent Head}
\label{sec:vdhead}
As described in \cref{sec:preliminaries}, the baseline pipeline reconstructs a static set of Gaussians in the canonical space. However, due to its fixed geometry and the inherent limitations of SH, this static formulation often struggles to capture complex view-dependent effects, such as specular highlights, reflections, and subtle appearance variations. To address this limitation, we introduce a DPT-based view-dependent head that adopts a hypernetwork architecture~\cite{ha2017hypernetworks} to predict per-pixel residual offsets for the Gaussian parameters. Instead of directly outputting offsets, this head takes feature tokens from the geometry transformer to capture rich geometric and material context, and generates pixel-wise parameters for a secondary network, which we refer to as a View MLP. 

This design allows the model to generate pixel-specific response functions tailored to the local properties of the scene. The generated View MLPs are conditioned on the target camera pose and compute the corresponding Gaussian parameter offsets. By predicting these View MLPs on the fly, the architecture can flexibly adapt its refinement strategy to the specific requirements of the target viewpoint. For instance, the head can generate a more complex mapping for specular surfaces to capture shifting reflections, while maintaining simpler functions for diffuse regions. This separation between context-driven weight generation and pose-driven offset calculation enables the model to capture intricate high-frequency appearance variations without being constrained by a static predefined parameter space.

\subsection{View-Dependent Gaussian Refinement}
\label{sec:vdrefine}
We now describe how the generated View MLPs utilize the target pose and refine the canonical Gaussians. The View MLP takes a 4D per-pixel target pose representation as input to compute Gaussian parameter offsets. This 4D pose representation, comprising a 3D unit viewing direction and a 1D distance term, is carefully derived from the target camera extrinsics $P^{t \to 1}$.

\subsubsection{Target Pose Reparameterization.}
The raw target extrinsic $P^{t \to 1} \in \mathbb{R}^{4 \times 4}$ is a homogeneous world-to-camera transformation matrix composed of a $3 \times 3$ rotation matrix and a 3D translation vector, as defined in \cref{eq:contextpose}. Directly feeding this full transformation matrix into the network is undesirable as the complex mathematical constraints inherent in transformation matrices (\eg, rotation orthogonality) are difficult for an MLP to optimize. To alleviate this, we reparameterize the target pose into a compact and geometrically meaningful representation consisting of a 3D unit viewing direction and a 1D distance. This design reduces the input dimensionality from 16 scalar elements to 4, while retaining essential view-dependent cues, facilitating network learning. We detail the formal mathematical derivation of the target pose representation in \cref{sec:target_pose_detail}.

\subsubsection{Gaussian Refinement.}
Given the 4D target pose representation, the View MLP predicts parameter offsets for the Gaussians at each pixel. Specifically, the network outputs the following offset values:
\begin{equation}
    \{\Delta \mathcal{G}^{v \to 1}\}_{v=1, \dots, N} = \{(\Delta \mu^{v \to 1 | t}_j, \Delta \alpha^{v | t}_j, \Delta r^{v \to 1 | t}_j, \Delta s^{v | t}_j, \Delta c^{v \to 1 | t}_j)\}_{j=1, \dots, H \times W}.
    \label{eq:offsetvalues}
\end{equation}
These offsets encode how each Gaussian, originating from a specific context view and pixel, should adapt when observed from the target viewpoint $t$. The predicted offsets are directly applied to the canonical 3D Gaussians defined in \cref{eq:gaussianparams} via element-wise residual addition (\eg, $\mathbf{\hat{\mu}} = \mu + \Delta \mu$), yielding the final refined parameters $\{\hat{\mathcal{G}}^{v \to 1} \}_{v=1, \dots, N}$. The rasterizer then renders the novel view from the target pose $P^{t \to 1}$:
\begin{equation}
    \hat{I}^t = \mathcal{R}(P^{t \to 1}, \{\hat{\mathcal{G}}^{v \to 1} \}_{v=1, \dots, N}).
    \label{eq:rendering}
\end{equation}

\subsection{Loss Function}
\label{sec:loss}
Following the training strategy of SPFSplat~\cite{huang2025spfsplat}, our model is optimized end-to-end without relying on ground-truth camera poses. The overall training objective consists of an image rendering loss $\mathcal{L}_{render}$ and a geometric reprojection loss $\mathcal{L}_{reproj}$. Specifically, $\mathcal{L}_{render}$ is formulated as a combination of MSE photometric loss and LPIPS~\cite{zhang2018lpips} perceptual loss (with a weight of $\lambda_{LPIPS}=0.05$) to balance low-level photometric accuracy and high-level perceptual similarity. To stabilize training in the canonical space and enforce explicit geometric constraints, we employ $\mathcal{L}_{reproj}$, which penalizes the distance between 2D pixel coordinates and the projected 3D Gaussian centers using estimated relative poses. The total loss is defined as:
\begin{equation}
    \mathcal{L}_{total} = \mathcal{L}_{render} + \lambda_{reproj}\mathcal{L}_{reproj},
    \label{eq:totalloss}
\end{equation}
where $\lambda_{reproj}$ is a weighting factor (set to $0.001$).
We provide full mathematical derivations and further implementation details in \cref{sec:loss_detail}.

\section{Experiments}
We evaluate the performance of our method on novel view synthesis and cross-dataset generalization across multiple datasets. In addition, we conduct comprehensive ablation studies to analyze the contribution of each component.

\subsection{Experimental Settings}
\subsubsection{Datasets.}
We train and evaluate our method on the RealEstate10K (RE10K)~\cite{zhou2018re10k}, which contains large-scale real estate videos, and ACID~\cite{liu2021acid}, which consists of nature scenes captured by aerial drones. The camera poses for both datasets are obtained via SfM, but our method operates in a strictly pose-free manner, entirely bypassing the need for such precomputed information.
We adopt the official train--test splits used in prior works~\cite{charatan2024pixelsplat,chen2024mvsplat,ye2025noposplat,huang2025spfsplat}.
Furthermore, to validate cross-dataset generalization, we follow~\cite{chen2024mvsplat,ye2025noposplat,huang2025spfsplat} and evaluate our model on ACID, the object-centric DTU dataset~\cite{jensen2014dtu}, and DL3DV~\cite{ling2024dl3dv}---a large-scale outdoor benchmark comprising 10K videos with complex camera trajectories that extend far beyond RE10K.

\subsubsection{Baselines.}
We compare our method with several baselines for novel view synthesis. These baselines include pose-required models (pixelSplat~\cite{charatan2024pixelsplat} and MVSplat~\cite{chen2024mvsplat}), supervised pose-free models (CoPoNeRF~\cite{hong2024coponerf}, Splatt3R~\cite{smart2024splatt3r}, and NoPoSplat~\cite{ye2025noposplat}), and self-supervised models (SelfSplat \cite{kang2025selfsplat}, PF3plat~\cite{hong2025pf3plat}, SPFSplat~\cite{huang2025spfsplat}, and SPFSplatV2~\cite{huang2025spfsplatv2}).

\subsubsection{Evaluation Protocols.}
We assess the quality of novel view synthesis using standard metrics: pixel-level PSNR, patch-level SSIM~\cite{wang2004ssim}, and feature-level LPIPS~\cite{zhang2018lpips}.

During NVS evaluation, target views are conventionally rendered using ground-truth poses~\cite{charatan2024pixelsplat,chen2024mvsplat,hong2024coponerf,smart2024splatt3r}. Alternative approaches include rendering with estimated target poses (\eg, PF3plat~\cite{hong2025pf3plat}, SelfSplat~\cite{kang2025selfsplat}, SPFSplat~\cite{huang2025spfsplat}, and SPFSplatV2~\cite{huang2025spfsplatv2}) or employing an evaluation-time pose alignment (EPA) strategy (NoPoSplat~\cite{ye2025noposplat}), which optimizes target-view poses during evaluation while freezing the reconstructed Gaussians. In our experiments, we evaluate our method using the estimated target poses. This setting allows us to jointly evaluate the reconstruction quality and the alignment between the estimated poses and the learned Gaussians.

\subsection{Implementation Details}
We implement ViewSplat in PyTorch, utilizing a CUDA-based 3DGS renderer with gradient support for camera poses. All models are trained on multiple NVIDIA RTX 4090 GPUs, while inference is performed on a single GPU.

Following prior works~\cite{huang2025spfsplat,huang2025spfsplatv2}, we adopt the geometry transformer backbones from MASt3R~\cite{leroy2024mast3r} for the SPFSplat and SPFSplatV2 architectures, and VGGT~\cite{wang2025vggt} for the V2-L variant.
To ensure stable convergence, we initialize the geometry transformer, Gaussian parameter head, Gaussian center head, and pose head using pretrained weights from the corresponding SPFSplat variants.
The remaining component, our view-dependent head, is zero-initialized; this allows the network to initially preserve the baseline static reconstruction while progressively learning to incorporate view-adaptive refinements.

The view-dependent head generates pixel-wise View MLPs with a single hidden layer of 16 dimensions. All experiments are conducted at a resolution of $256 \times 256$, except for the V2-L variant, which follows its backbone's default resolution of $224 \times 224$.

\subsubsection{Training Strategy.}
We adopt a frozen-backbone training strategy, where the pretrained weights of the base models (SPFSplat or SPFSplatV2) remain fixed while only the view-dependent head is optimized. We use the Adam optimizer with a learning rate of $1 \times 10^{-4}$ and a batch size of 12. Each batch contains a single scene with its corresponding input and target views. Consistent with prior work, the frame distance between input views is progressively increased during training to facilitate curriculum learning.

\begin{table}[t]
\centering
\caption{Performance comparison of novel view synthesis on RE10K~\cite{zhou2018re10k} and ACID~\cite{liu2021acid} datasets. We report the average metrics across all scenes. Results using the \textbf{ViewSplat} framework are highlighted in \textbf{bold}, demonstrating consistent performance gains across all SPFSplat variants. * indicates the use of the evaluation-time pose alignment (EPA) strategy.}
\label{tab:nvs_main_average}
\setlength{\tabcolsep}{5pt}
\begin{scriptsize}
    \begin{tabular}{l ccc ccc}
        \toprule
        \multirow{2}{*}{\textbf{Method}} & \multicolumn{3}{c}{\textbf{RE10K}} & \multicolumn{3}{c}{\textbf{ACID}} \\
            & PSNR$\uparrow$ & SSIM$\uparrow$ & LPIPS$\downarrow$ & PSNR$\uparrow$ & SSIM$\uparrow$ & LPIPS$\downarrow$ \\ \midrule
        \textit{Pose-Required} & & & & & & \\
            pixelSplat & 23.859 & 0.808 & 0.184 & 25.889 & 0.780 & 0.194 \\
            MVSplat & 24.012 & 0.812 & 0.175 & 25.561 & 0.775 & 0.195 \\ \midrule
        \textit{Supervised Pose-Free} & & & & & & \\
            CoPoNeRF & 18.938 & 0.619 & 0.388 & 20.950 & 0.606 & 0.406 \\
            Splatt3R & 18.688 & 0.337 & 0.596 & 18.060 & 0.510 & 0.407 \\
            NoPoSplat* & 25.033 & 0.838 & 0.160 & 25.961 & 0.781 & 0.189 \\ \midrule
        \textit{Self-Supervised Pose-Free} & & & & & & \\
            SelfSplat & 19.152 & 0.680 & 0.328 & 22.089 & 0.694 & 0.298 \\
            PF3plat & 21.042 & 0.739 & 0.233 & 21.206 & 0.632 & 0.293 \\ \midrule
            SPFSplat & 25.484 & 0.847 & 0.153 & 26.070 & 0.781 & 0.186 \\
            \textbf{+ViewSplat} & \textbf{26.317} & \textbf{0.857} & \textbf{0.144} & \textbf{26.661} & \textbf{0.790} & \textbf{0.177} \\ \midrule
            SPFSplatV2 & 25.693 & 0.853 & 0.149 & 26.284 & 0.791 & 0.182 \\
            \textbf{+ViewSplat} & \textbf{26.468} & \textbf{0.861} & \textbf{0.140} & \textbf{26.873} & \textbf{0.801} & \textbf{0.173} \\ \midrule
            SPFSplatV2-L & 25.668 & 0.855 & 0.137 & 26.674 & 0.806 & 0.162 \\
            \textbf{+ViewSplat}  & \textbf{26.798} & \textbf{0.870} & \textbf{0.124} & \textbf{27.509} & \textbf{0.820} & \textbf{0.149} \\ \bottomrule
    \end{tabular}
\end{scriptsize}
\end{table}

\subsection{Results}
\subsubsection{Novel View Synthesis.}
\begin{figure}[t]
    \centering
    \includegraphics[width=1\linewidth]{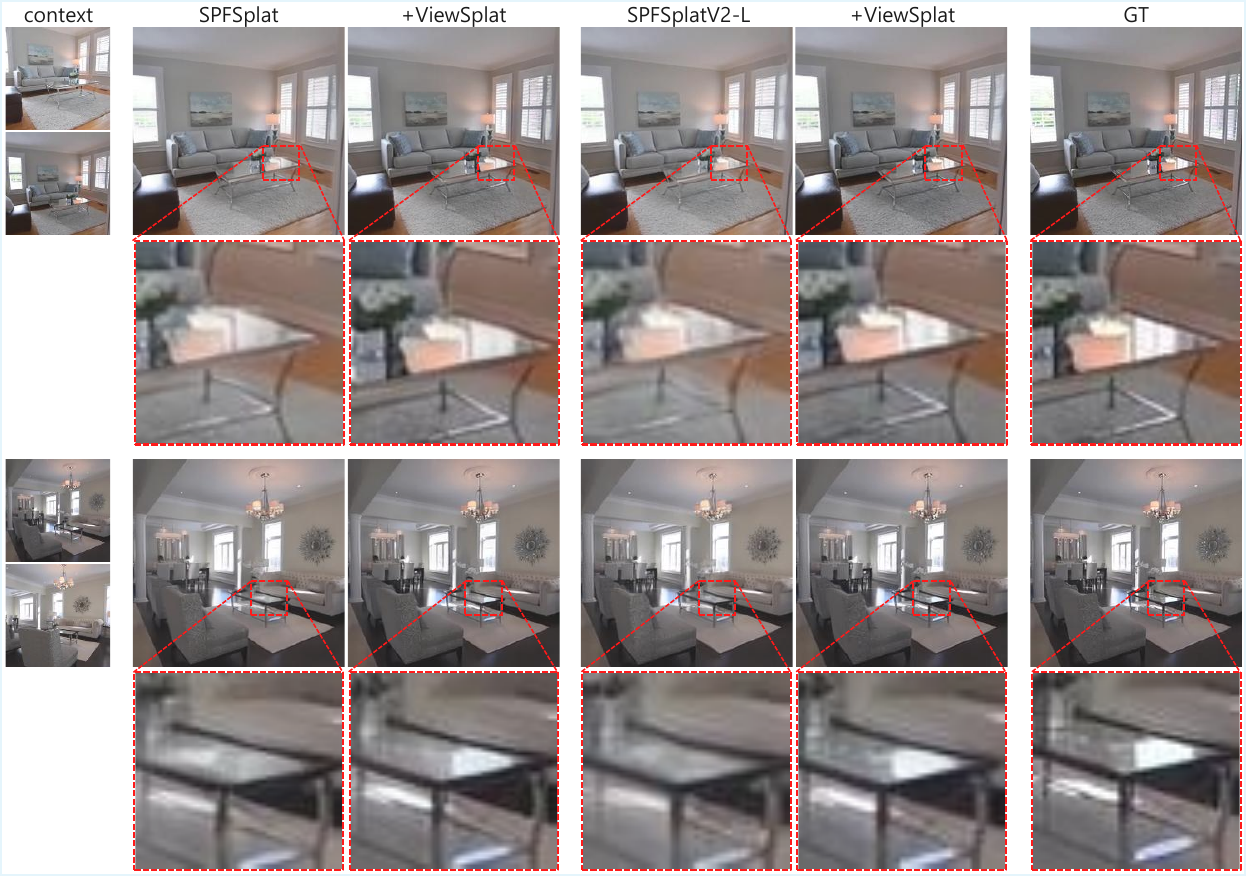}
    \caption{Qualitative comparison on RE10K. Compared to baseline methods, our approach accurately reconstructs sharp reflections.}
    \label{fig:qualitative_main_nvs}
\end{figure}

We evaluate ViewSplat on RE10K and ACID, and the quantitative results are summarized in \cref{tab:nvs_main_average}. Our framework, particularly the V2-L variant, consistently achieves new state-of-the-art performance across all metrics.
A key finding is the orthogonal performance gain provided by our view-adaptive refinement. Across all backbone variants---SPFSplat, V2, and V2-L---ViewSplat consistently yields substantial improvements in rendering fidelity.
This consistency underscores the scalability and robustness of our approach in capturing intricate view-dependent details across diverse environments, from structured interiors to complex natural landscapes. Qualitative comparisons in \cref{fig:qualitative_main_nvs} further demonstrate ViewSplat's ability to synthesize high-frequency specularities and maintain geometric consistency, resulting in photorealistic novel views.
We provide detailed results with varying levels of camera overlap and additional qualitative results in \cref{sec:camera_overlap_detail} and \cref{sec:more_qualitative}.

\begin{table}
    \centering

\begin{minipage}{0.48\textwidth}
    \centering
    \caption{Comparison of efficiency on a single NVIDIA RTX 4090 GPU using RE10K. ViewSplat is evaluated on three SPFSplat backbone variants.}
    \label{tab:efficiency}
    \setlength{\tabcolsep}{3pt}
    \resizebox{\linewidth}{!}{
        \begin{tabular}{l ccc}
            \toprule
            \multirow{2}{*}{\textbf{Method}} & \textbf{Params}$\downarrow$ & \textbf{Inference time}$\downarrow$ & \textbf{Rendering speed}$\uparrow$ \\
                & (B) & (s) & (FPS) \\ \midrule
            SPFSplat & 0.616 & 0.046 & 386 \\
            \textbf{+ViewSplat} & 0.658 & 0.057 & 154 \\ \midrule
            SPFSplatV2 & 0.613 & 0.039 & 390 \\
            \textbf{+ViewSplat} & 0.655 & 0.052 & 149 \\ \midrule
            SPFSplatV2-L & 1.223 & 0.066 & 231 \\
            \textbf{+ViewSplat} & 1.256 & 0.068 & 90 \\ \bottomrule
        \end{tabular}
    }
\end{minipage}
\hfill
\begin{minipage}{0.48\textwidth}
    \centering
    \caption{Comparison of rendering performance between SPFSplat with varying SH degrees and the proposed ViewSplat on RE10K. ``SPFSplat (Std.)'' denotes the standard SPFSplat setting with SH degree 4. The \textbf{best} results are highlighted.}
    \label{tab:sh_degree}
    \setlength{\tabcolsep}{4pt}
    \resizebox{\linewidth}{!}{
        \begin{tabular}{l c ccc}
            \toprule
            \textbf{Method} & \textbf{SH Degree} & PSNR$\uparrow$ & SSIM$\uparrow$ & LPIPS$\downarrow$ \\ \midrule
            SPFSplat (Std.) & 4 & 25.486 & 0.847 & 0.153 \\
            SPFSplat & 6 & 25.408 & 0.845 & 0.154 \\
            SPFSplat & 8 & 25.418 & 0.845 & 0.153 \\ \midrule
            \textbf{+ViewSplat} & 4 & \textbf{26.317} & \textbf{0.857} & \textbf{0.144} \\ \bottomrule
        \end{tabular}
    }
\end{minipage}
\end{table}

\subsubsection{Efficiency.}
We evaluate the computational efficiency of our framework on a single NVIDIA RTX 4090 GPU (\cref{tab:efficiency}). 
Integrating the ViewSplat module introduces minimal overhead to the forward pass, adding only 11--13~ms to jointly reconstruct the 3D Gaussians and predict the View MLP parameters. 

However, target-pose conditioning incurs a substantial drop in rendering speed (\eg, from 386 to 154~FPS for the base backbone) since residual attributes must be re-evaluated for each novel viewpoint. 
Nevertheless, even with our largest SPFSplatV2-L backbone, ViewSplat comfortably maintains 90~FPS, which is well above the practical real-time threshold ($\geq$ 30~FPS). 
Thus, while the rendering-time overhead is non-negligible, our framework successfully balances enhanced view-dependent fidelity with interactive rendering speeds.

\subsubsection{Analysis of View-Dependent Modeling.}
To further validate the effectiveness of our view-dependent refinement framework, we compare ViewSplat against the baseline SPFSplat with varying SH degrees (\cref{tab:sh_degree}). While increasing the SH degree of the baseline model from 4 to 8 yields negligible improvement and may even slightly degrade performance, ViewSplat with SH degree 4 significantly outperforms all baseline variants. This suggests that standard SH-based representations reach a performance limit in pose-free settings, whereas our view-dependent refinement provides superior expressive capacity for capturing complex specularities and non-linear appearances.

\begin{table}
    \centering
\caption{\textbf{Cross-dataset generalization.} All methods are trained on RE10K and evaluated in a zero-shot setting on ACID, DTU, and DL3DV. Results using the \textbf{ViewSplat} framework are highlighted in \textbf{bold}, demonstrating consistent performance gains across all SPFSplat variants.}
\label{tab:cross-dataset}
\begin{scriptsize}
    \resizebox{\linewidth}{!}{%
        \begin{tabular}{l ccc ccc ccc}
            \toprule
            \multirow{2}{*}{\textbf{Method}} & \multicolumn{3}{c}{\textbf{ACID}} & \multicolumn{3}{c}{\textbf{DTU}} & \multicolumn{3}{c}{\textbf{DL3DV}} \\
                & PSNR$\uparrow$ & SSIM$\uparrow$ & LPIPS$\downarrow$ & PSNR$\uparrow$ & SSIM$\uparrow$ & LPIPS$\downarrow$ & PSNR$\uparrow$ & SSIM$\uparrow$ & LPIPS$\downarrow$ \\ \midrule
            \textit{Pose-Required} & & & & & & \\
            pixelSplat & 25.477 & 0.770 & 0.207 & 15.067 & 0.539 & 0.341 & 18.688 & 0.582 & 0.354 \\
            MVSplat & 25.525 & 0.773 & 0.199 & 14.542 & 0.537 & 0.324 & 17.786 & 0.545 & 0.357 \\ \midrule
            \textit{Supervised Pose-Free} & & & & & & \\
            NoPoSplat* & 25.764 & 0.776 & 0.199 & 17.899 & 0.629 & 0.279 & 19.974 & 0.612 & 0.305 \\ \midrule
            \textit{Self-Supervised Pose-Free} & & & & & & \\
            SelfSplat & 22.204 & 0.686 & 0.316 & 13.249 & 0.434 & 0.441 & 15.047 & 0.410 & 0.498 \\
            PF3plat & 20.726 & 0.610 & 0.308 & 12.972 & 0.407 & 0.464 & 15.773 & 0.458 & 0.417 \\ \midrule
            SPFSplat & 25.965 & 0.781 & 0.190 & 16.550 & 0.579 & 0.270 & 19.172 & 0.573 & 0.315 \\
            \textbf{+ViewSplat} & \textbf{26.595} & \textbf{0.793} & \textbf{0.179} & \textbf{16.864} & \textbf{0.589} & \textbf{0.259} & \textbf{19.858} & \textbf{0.591} & \textbf{0.298} \\ \midrule
            SPFSplatV2 & 26.220 & 0.789 & 0.185 & 16.793 & 0.584 & 0.265 & 19.439 & 0.584 & 0.304 \\
            \textbf{+ViewSplat} & \textbf{26.761} & \textbf{0.798} & \textbf{0.175} & \textbf{17.034} & \textbf{0.588} & \textbf{0.258} & \textbf{19.953} & \textbf{0.599} & \textbf{0.290} \\ \midrule
            SPFSplatV2-L & 26.361 & 0.796 & 0.169 & 17.739 & 0.653 & 0.228 & 19.743 & 0.613 & 0.277 \\
            \textbf{+ViewSplat} & \textbf{27.169} & \textbf{0.810} & \textbf{0.154} & \textbf{18.174} & \textbf{0.665} & \textbf{0.215} & \textbf{20.563} & \textbf{0.635} & \textbf{0.256} \\ \bottomrule
        \end{tabular}
    }
\end{scriptsize}
\end{table}

\begin{figure}[t]
    \centering
    \includegraphics[width=1\linewidth]{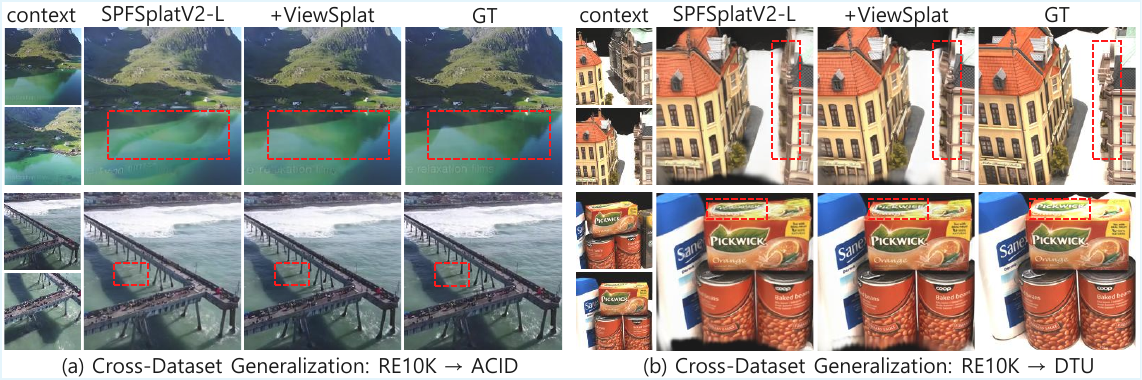}
    \caption{Qualitative results of cross-dataset generalization.}
    \label{fig:qualitative_cross_dataset}
\end{figure}

\subsubsection{Cross-Dataset Generalization.}
To evaluate the generalization capability of our method, we conduct zero-shot evaluations on the ACID~\cite{liu2021acid}, DTU~\cite{jensen2014dtu}, and DL3DV~\cite{ling2024dl3dv} datasets after training exclusively on RE10K. DTU and DL3DV are particularly challenging out-of-distribution settings for RE10K-trained models, due to their object-centric scenes and large-scale outdoor trajectories, respectively, which helps explain their lower absolute PSNR values. As shown in \cref{tab:cross-dataset} and \cref{fig:qualitative_cross_dataset}, ViewSplat consistently improves all SPFSplat variants across datasets, with the strongest performance obtained using the SPFSplatV2-L backbone. These results indicate that our view-dependent refinement learns transferable representations that generalize beyond the RE10K training distribution.

\subsection{Ablation Studies}
\subsubsection{Effectiveness of Hypernetwork-style Formulation vs. Direct Regression.}
We compare our hypernetwork-style scene-conditioned View MLP method with a \emph{Direct Regression} baseline that predicts pixel-wise Gaussian offsets from concatenated context features and a target pose (\cref{tab:direct_offset}). Direct regression yields a small PSNR gain over ours (+0.08 dB), but reduces rendering speed from 154 to 72 FPS, corresponding to a \textbf{+436\%} rendering-time overhead over SPFSplat. This is because the target pose is coupled with the heavy DPT decoder, requiring the dense network to be rerun for each target-view query.

In contrast, our hypernetwork formulation predicts per-pixel View MLP weights \emph{once} during inference and applies only lightweight pose-conditioned View MLPs during rendering. Thus, ViewSplat preserves real-time rendering while achieving fidelity comparable to direct regression.

\begin{table}
    \centering
\caption{\textbf{Architectural design analysis on RE10K.} We compare our hypernetwork-style scene-conditioned View MLP with a direct offset regression baseline.}
\label{tab:direct_offset}

\setlength{\tabcolsep}{3pt}
\begin{scriptsize}
    \begin{tabular}{l ccc | cc}
        \toprule
        \textbf{Method} & PSNR $\uparrow$ & SSIM $\uparrow$ & LPIPS $\downarrow$ & Inference Time (s) $\downarrow$ & Rendering Speed (FPS) $\uparrow$ \\ \midrule
            SPFSplat & 25.484 & 0.847 & 0.153 & \textbf{0.046} & \textbf{386} \\
            View MLP (Ours) & 26.317 & 0.857 & 0.144 & 0.057 & 154 \\ 
            Direct Regression & \textbf{26.396} & \textbf{0.858} & \textbf{0.143} & 0.048 & 72 \\ \bottomrule
    \end{tabular}
\end{scriptsize}
\end{table}

\subsubsection{Ablation on Refinement Components.}
\begin{figure}
    \centering
\begin{minipage}[c]{0.73\textwidth}
    \centering
    \captionof{table}{\textbf{Effectiveness of refinement components.} We evaluate the contribution of residual offsets for each Gaussian parameter ($\Delta\mu$, $\Delta\alpha$, $\Delta r$, $\Delta s$, and $\Delta c$) on RE10K. The \textbf{best} and \underline{second-best} results are highlighted.}
    \label{tab:ablation_offsets}
\begin{scriptsize}
    \resizebox{\textwidth}{!}{
        \begin{tabular}{l cccc | ccc}
            \toprule
            \textbf{Method} & $\Delta\mu$ & $\Delta\alpha$ & $\Delta(r,s)$ & $\Delta c$ & PSNR$\uparrow$ & SSIM$\uparrow$ & LPIPS$\downarrow$ \\ \midrule
            (1) Baseline (SPFSplat) & $\times$ & $\times$ & $\times$ & $\times$ & 25.484 & 0.847 & 0.153 \\
            (2) $\Delta(\mu,\alpha)$ only & $\checkmark$ & $\checkmark$ & $\times$ & $\times$ & 25.760 & 0.850 & 0.155 \\
            (3) $\Delta(r,s,c)$ only & $\times$ & $\times$ & $\checkmark$ & $\checkmark$ & 26.241 & \underline{0.856} & 0.146 \\
            (4) w/o $\Delta\alpha$ & $\checkmark$ & $\times$ & $\checkmark$ & $\checkmark$ & 15.684 & 0.555 & 0.525 \\
            (5) w/o $\Delta\mu$ & $\times$ & $\checkmark$ & $\checkmark$ & $\checkmark$ & \underline{26.307} & \textbf{0.857} & \underline{0.145} \\
            (6) w/o $\Delta(r,s)$ & $\checkmark$ & $\checkmark$ & $\times$ & $\checkmark$ & 26.282 & \textbf{0.857} & \underline{0.145} \\
            (7) w/o $\Delta c$ & $\checkmark$ & $\checkmark$ & $\checkmark$ & $\times$ & 26.098 & \underline{0.856} & \underline{0.145} \\
            \textbf{(8) Full (ViewSplat)} & $\checkmark$ & $\checkmark$ & $\checkmark$ & $\checkmark$ & \textbf{26.317} & \textbf{0.857} & \textbf{0.144} \\ \bottomrule
        \end{tabular}
    }
\end{scriptsize}
\end{minipage}
\hfill
\begin{minipage}[c]{0.25\textwidth}
    \centering
    \includegraphics[width=\textwidth]{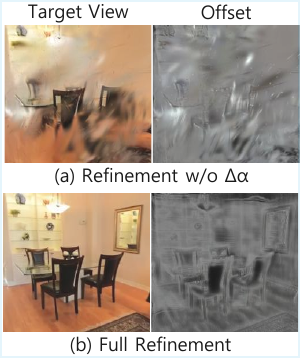}
    \captionof{figure}{Artifacts from decoupled $\mu/\alpha$ refinement.}
    \label{fig:qualitative_mrsc}
\end{minipage}
\end{figure}

\cref{tab:ablation_offsets} shows that most refinement subsets improve performance over the baseline. However, case (4) suffers from catastrophic degradation when mean ($\mu$) offsets are applied without corresponding opacity refinement. This decoupling leads to severe artifacts, blurring, and structural collapse, as shown in \cref{fig:qualitative_mrsc}. These results confirm that spatial shifts must be synchronized with visibility updates to preserve structural coherence. By jointly optimizing all parameters, our full model (8) achieves the best performance across all metrics, underscoring the importance of holistic refinement.

\subsubsection{Effect of View MLP Hidden Dimension.}
As shown in \cref{tab:ablation_hiddendim}, reducing the hidden dimension $D$ of the View MLP improves rendering speed and memory efficiency with minimal quality loss. Decreasing $D$ from 16 to 8 increases rendering speed from 154 to 186 FPS and reduces memory usage by 18.5\%, while PSNR drops by only 0.003 dB. Further reducing $D$ to 4 provides additional speedup but leads to a larger quality drop. Thus, while we use $D=16$ in the main experiments for maximum fidelity, $D=8$ offers the best trade-off between efficiency and quality. 
Additional analyses, including backbone generalization with YoNoSplat~\cite{ye2026yonosplat} and context-view scaling, are provided in \cref{sec:additional_exp}.

\begin{table}
    \centering
\caption{Quantitative evaluation of rendering quality and efficiency on RE10K with varying hidden dimensions ($D$) of the View MLP in our ViewSplat framework.}
\label{tab:ablation_hiddendim}

\setlength{\tabcolsep}{5pt}
\begin{scriptsize}
    \begin{tabular}{l ccc c c c}
        \toprule
        \textbf{D} & PSNR$\uparrow$ & SSIM$\uparrow$ & LPIPS$\downarrow$ & \textbf{Inf. Time (s)}$\downarrow$ & \textbf{FPS}$\uparrow$ & \textbf{Mem. (GiB)}$\downarrow$ \\ \midrule
        16 & 26.317 & 0.857 & 0.144 & 0.057 & 154 & 0.648 \\
        8 & 26.314 & 0.857 & 0.144 & 0.057 & 186 & 0.528 \\
        4 & 26.302 & 0.857 & 0.144 & 0.056 & 207 & 0.487 \\ \bottomrule
    \end{tabular}
\end{scriptsize}
\end{table}

\subsection{Limitations}
\begin{figure}
    \centering
    \includegraphics[width=1\linewidth]{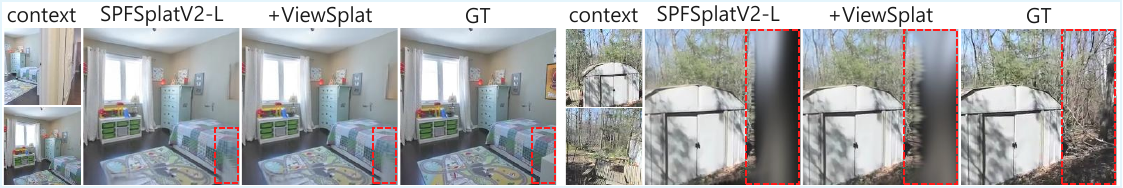}
    \caption{Limitations: visual artifacts in unobserved regions.}
    \label{fig:qualitative_fail}
\end{figure}

Despite the improvements achieved by ViewSplat through its view-dependent refinement framework, it inherits a key limitation of the baseline SPFSplat: the inability to synthesize content in unobserved regions. Since both methods are reconstruction-based and lack generative priors, they cannot hallucinate missing information, which results in blurry artifacts or empty regions in sparse-view areas, as illustrated in \cref{fig:qualitative_fail}. This limitation highlights the inherent challenge of achieving complete scene recovery without incorporating generative modeling.

Furthermore, our view-adaptive design introduces an inherent trade-off between rendering quality and computational throughput. Because the Gaussian attributes are refined on the fly, conditioned on each specific target-view pose, the attributes must be re-evaluated for every novel viewpoint. This target-view pose conditioning prevents the framework from caching a single static 3D scene representation, leading to a substantial reduction in rendering frame rates compared to the static baseline (as quantified in \cref{tab:efficiency}). It also implies that multi-view consistency is maintained within a family of view-conditioned approximations rather than a single rigid geometry, which remains an important direction for future investigation.

\section{Conclusion}
We present ViewSplat, a view-adaptive framework that addresses the fidelity bottleneck in feed-forward 3D Gaussian splatting. By shifting from static primitive regression to view-adaptive splatting, our method refines Gaussian attributes conditioned on each target view, enabling the modeling of complex view-dependent appearance that rigid representations often fail to capture. ViewSplat achieves state-of-the-art fidelity while maintaining real-time rendering (90--154 FPS) across multiple benchmarks, without requiring precomputed poses.
We hope our view-adaptive paradigm inspires future research into more flexible and high-fidelity 3D representations for real-world scenarios.


\section*{Acknowledgements}
This work was supported by the National Research Foundation of Korea(NRF) grant funded by the Korea government(MSIT) (RS-2026-25481939) and the Industrial Core Technology Development Program of MOTIE/KEIT (RS-2025-02307680, Development of an inline vision inspection system based on the on-device AI semiconductor for real-time large-area defect detection). The authors acknowledge the Urban Big data and AI Institute of the University of Seoul supercomputing resources (http://ubai.uos.ac.kr) made available for conducting the research reported in this paper.

%
%
\bibliographystyle{splncs04}
\bibliography{main}

\begin{thebibliography}{10}
\providecommand{\url}[1]{\texttt{#1}}
\providecommand{\urlprefix}{URL }
\providecommand{\doi}[1]{https://doi.org/#1}

\bibitem{boss2021nerd}
Boss, M., Braun, R., Jampani, V., Barron, J.T., Liu, C., Lensch, H.P.: Nerd: Neural reflectance decomposition from image collections. In: Int. Conf. Comput. Vis. pp. 12684--12694 (2021)

\bibitem{boss2021neural-pil}
Boss, M., Jampani, V., Braun, R., Liu, C., Barron, J., Lensch, H.P.: Neural-pil: Neural pre-integrated lighting for reflectance decomposition. In: Adv. Neural Inform. Process. Syst. pp. 10691--10704 (2021)

\bibitem{charatan2024pixelsplat}
Charatan, D., Li, S.L., Tagliasacchi, A., Sitzmann, V.: pixelsplat: 3d gaussian splats from image pairs for scalable generalizable 3d reconstruction. In: IEEE Conf. Comput. Vis. Pattern Recog. pp. 19457--19467 (2024)

\bibitem{chen2024marepo}
Chen, S., Cavallari, T., Prisacariu, V.A., Brachmann, E.: Map-relative pose regression for visual re-localization. In: IEEE Conf. Comput. Vis. Pattern Recog. pp. 20665--20674 (2024)

\bibitem{chen2024mvsplat}
Chen, Y., Xu, H., Zheng, C., Zhuang, B., Pollefeys, M., Geiger, A., Cham, T.J., Cai, J.: Mvsplat: Efficient 3d gaussian splatting from sparse multi-view images. In: Eur. Conf. Comput. Vis. pp. 370--386 (2024)

\bibitem{dosovitskiy2021vit}
Dosovitskiy, A., Beyer, L., Kolesnikov, A., Weissenborn, D., Zhai, X., Unterthiner, T., Dehghani, M., Minderer, M., Heigold, G., Gelly, S., Uszkoreit, J., Houlsby, N.: An image is worth 16x16 words: Transformers for image recognition at scale. In: Int. Conf. Learn. Represent. (2021)

\bibitem{edstedt2024roma}
Edstedt, J., Sun, Q., B\"okman, G., Wadenb\"ack, M., Felsberg, M.: Roma: Robust dense feature matching. In: IEEE Conf. Comput. Vis. Pattern Recog. pp. 19790--19800 (2024)

\bibitem{ha2017hypernetworks}
Ha, D., Dai, A.M., Le, Q.V.: Hypernetworks. In: Int. Conf. Learn. Represent. (2017)

\bibitem{hong2025pf3plat}
Hong, S., Jung, J., Shin, H., Han, J., Yang, J., Luo, C., Kim, S.: Pf3plat: Pose-free feed-forward 3d gaussian splatting. In: Int. Conf. Mach. Learn. (2025)

\bibitem{hong2024coponerf}
Hong, S., Jung, J., Shin, H., Yang, J., Kim, S., Luo, C.: Unifying correspondence pose and nerf for generalized pose-free novel view synthesis. In: IEEE Conf. Comput. Vis. Pattern Recog. pp. 20196--20206 (2024)

\bibitem{huang2025spfsplat}
Huang, R., Mikolajczyk, K.: No pose at all: Self-supervised pose-free 3d gaussian splatting from sparse views. In: Int. Conf. Comput. Vis. pp. 27947--27957 (2025)

\bibitem{huang2025spfsplatv2}
Huang, R., Mikolajczyk, K.: Spfsplatv2: Efficient self-supervised pose-free 3d gaussian splatting from sparse views (2025), arXiv preprint arXiv:2509.17246

\bibitem{jensen2014dtu}
Jensen, R., Dahl, A., Vogiatzis, G., Tola, E., Aanaes, H.: Large scale multi-view stereopsis evaluation. In: IEEE Conf. Comput. Vis. Pattern Recog. pp. 406--413 (2014)

\bibitem{jiang2025anysplat}
Jiang, L., Mao, Y., Xu, L., Lu, T., Ren, K., Jin, Y., Xu, X., Yu, M., Pang, J., Zhao, F., et~al.: Anysplat: Feed-forward 3d gaussian splatting from unconstrained views. ACM Trans. Graph.  \textbf{44}(6),  1--16 (2025)

\bibitem{jiang2024gaussianshader}
Jiang, Y., Tu, J., Liu, Y., Gao, X., Long, X., Wang, W., Ma, Y.: Gaussianshader: 3d gaussian splatting with shading functions for reflective surfaces. In: IEEE Conf. Comput. Vis. Pattern Recog. pp. 5322--5332 (2024)

\bibitem{kang2025selfsplat}
Kang, G., Yoo, J., Park, J., Nam, S., Im, H., Shin, S., Kim, S., Park, E.: Selfsplat: Pose-free and 3d prior-free generalizable 3d gaussian splatting. In: IEEE Conf. Comput. Vis. Pattern Recog. pp. 22012--22022 (2025)

\bibitem{kerbl20233dgs}
Kerbl, B., Kopanas, G., Leimk{\"u}hler, T., Drettakis, G.: 3d gaussian splatting for real-time radiance field rendering. ACM Trans. Graph.  \textbf{42}(4),  1--14 (2023)

\bibitem{leroy2024mast3r}
Leroy, V., Cabon, Y., Revaud, J.: Grounding image matching in 3d with mast3r. In: Eur. Conf. Comput. Vis. pp. 71--91 (2024)

\bibitem{liang2023envidr}
Liang, R., Chen, H., Li, C., Chen, F., Panneer, S., Vijaykumar, N.: Envidr: Implicit differentiable renderer with neural environment lighting. In: Int. Conf. Comput. Vis. pp. 79--89 (2023)

\bibitem{lin2026da3}
Lin, H., Chen, S., Liew, J.H., Chen, D.Y., Li, Z., Zhao, Y., Peng, S., Guo, H., Zhou, X., Shi, G., Feng, J., Kang, B.: Depth anything 3: Recovering the visual space from any views. In: Int. Conf. Learn. Represent. (2026)

\bibitem{ling2024dl3dv}
Ling, L., Sheng, Y., Tu, Z., Zhao, W., Xin, C., Wan, K., Yu, L., Guo, Q., Yu, Z., Lu, Y., Li, X., Sun, X., Ashok, R., Mukherjee, A., Kang, H., Kong, X., Hua, G., Zhang, T., Benes, B., Bera, A.: Dl3dv-10k: A large-scale scene dataset for deep learning-based 3d vision. In: IEEE Conf. Comput. Vis. Pattern Recog. pp. 22160--22169 (2024)

\bibitem{liu2021acid}
Liu, A., Tucker, R., Jampani, V., Makadia, A., Snavely, N., Kanazawa, A.: Infinite nature: Perpetual view generation of natural scenes from a single image. In: Int. Conf. Comput. Vis. pp. 14458--14467 (2021)

\bibitem{liu2024mirrorgaussian}
Liu, J., Tang, X., Cheng, F., Yang, R., Li, Z., Liu, J., Huang, Y., Lin, J., Liu, S., Wu, X., et~al.: Mirrorgaussian: Reflecting 3d gaussians for reconstructing mirror reflections. In: Eur. Conf. Comput. Vis. pp. 377--393 (2024)

\bibitem{liu2023nero}
Liu, Y., Wang, P., Lin, C., Long, X., Wang, J., Liu, L., Komura, T., Wang, W.: Nero: Neural geometry and brdf reconstruction of reflective objects from multiview images. ACM Trans. Graph.  \textbf{42}(4),  1--22 (2023)

\bibitem{lu2024scaffold-gs}
Lu, T., Yu, M., Xu, L., Xiangli, Y., Wang, L., Lin, D., Dai, B.: Scaffold-gs: Structured 3d gaussians for view-adaptive rendering. In: IEEE Conf. Comput. Vis. Pattern Recog. pp. 20654--20664 (2024)

\bibitem{mildenhall2020nerf}
Mildenhall, B., Srinivasan, P.P., Tancik, M., Barron, J.T., Ramamoorthi, R., Ng, R.: Nerf: Representing scenes as neural radiance fields for view synthesis. Communications of the ACM  \textbf{65}(1),  99--106 (2021)

\bibitem{ranftl2021dpt}
Ranftl, R., Bochkovskiy, A., Koltun, V.: Vision transformers for dense prediction. In: Int. Conf. Comput. Vis. pp. 12179--12188 (2021)

\bibitem{schonberger2016sfmrevisited}
Schonberger, J.L., Frahm, J.M.: Structure-from-motion revisited. In: IEEE Conf. Comput. Vis. Pattern Recog. pp. 4104--4113 (2016)

\bibitem{smart2024splatt3r}
Smart, B., Zheng, C., Laina, I., Prisacariu, V.A.: Splatt3r: Zero-shot gaussian splatting from uncalibrated image pairs (2024), arXiv preprint arXiv:2408.13912

\bibitem{srinivasan2021nerv}
Srinivasan, P.P., Deng, B., Zhang, X., Tancik, M., Mildenhall, B., Barron, J.T.: Nerv: Neural reflectance and visibility fields for relighting and view synthesis. In: IEEE Conf. Comput. Vis. Pattern Recog. pp. 7495--7504 (2021)

\bibitem{tang2024lgm}
Tang, J., Chen, Z., Chen, X., Wang, T., Zeng, G., Liu, Z.: Lgm: Large multi-view gaussian model for high-resolution 3d content creation. In: Eur. Conf. Comput. Vis. pp. 1--18 (2024)

\bibitem{verbin2022ref-nerf}
Verbin, D., Hedman, P., Mildenhall, B., Zickler, T., Barron, J.T., Srinivasan, P.P.: Ref-nerf: Structured view-dependent appearance for neural radiance fields. In: IEEE Conf. Comput. Vis. Pattern Recog. pp. 5481--5490 (2022)

\bibitem{wang2025vggt}
Wang, J., Chen, M., Karaev, N., Vedaldi, A., Rupprecht, C., Novotny, D.: Vggt: Visual geometry grounded transformer. In: IEEE Conf. Comput. Vis. Pattern Recog. pp. 5294--5306 (2025)

\bibitem{wang2024dust3r}
Wang, S., Leroy, V., Cabon, Y., Chidlovskii, B., Revaud, J.: Dust3r: Geometric 3d vision made easy. In: IEEE Conf. Comput. Vis. Pattern Recog. pp. 20697--20709 (2024)

\bibitem{wang2004ssim}
Wang, Z., Bovik, A., Sheikh, H., Simoncelli, E.: Image quality assessment: from error visibility to structural similarity. IEEE Trans. Image Process.  \textbf{13}(4),  600--612 (2004)

\bibitem{xu2024grm}
Xu, Y., Shi, Z., Yifan, W., Chen, H., Yang, C., Peng, S., Shen, Y., Wetzstein, G.: Grm: Large gaussian reconstruction model for efficient 3d reconstruction and generation. In: Eur. Conf. Comput. Vis. pp. 1--20 (2024)

\bibitem{yang2024spec-gaussian}
Yang, Z., Gao, X., Sun, Y.T., Huang, Y.H., Lyu, X., Zhou, W., Jiao, S., Qi, X., Jin, X.: Spec-gaussian: Anisotropic view-dependent appearance for 3d gaussian splatting. In: Adv. Neural Inform. Process. Syst. pp. 61192--61216 (2024)

\bibitem{ye2026yonosplat}
Ye, B., Chen, B., Xu, H., Barath, D., Pollefeys, M.: Yonosplat: You only need one model for feedforward 3d gaussian splatting. In: Int. Conf. Learn. Represent. (2026)

\bibitem{ye2025noposplat}
Ye, B., Liu, S., Xu, H., Xueting, L., Pollefeys, M., Yang, M.H., Songyou, P.: No pose, no problem: Surprisingly simple 3d gaussian splats from sparse unposed images. In: Int. Conf. Learn. Represent. (2025)

\bibitem{zhang2024gs-lrm}
Zhang, K., Bi, S., Tan, H., Xiangli, Y., Zhao, N., Sunkavalli, K., Xu, Z.: Gs-lrm: Large reconstruction model for 3d gaussian splatting. In: Eur. Conf. Comput. Vis. pp. 1--19 (2024)

\bibitem{zhang2018lpips}
Zhang, R., Isola, P., Efros, A.A., Shechtman, E., Wang, O.: The unreasonable effectiveness of deep features as a perceptual metric. In: IEEE Conf. Comput. Vis. Pattern Recog. pp. 586--595 (2018)

\bibitem{Zhang2021nerfactor}
Zhang, X., Srinivasan, P.P., Deng, B., Debevec, P., Freeman, W.T., Barron, J.T.: Nerfactor: Neural factorization of shape and reflectance under an unknown illumination. ACM Trans. Graph.  \textbf{40}(6),  1--18 (2021)

\bibitem{zhou2018re10k}
Zhou, T., Tucker, R., Flynn, J., Fyffe, G., Snavely, N.: Stereo magnification: Learning view synthesis using multiplane images. ACM Trans. Graph.  \textbf{37}(4),  1--12 (2018)

\bibitem{zhou2025rtr-gs}
Zhou, Y., Zhang, F., Wang, Z., Zhang, L.: Rtr-gs: 3d gaussian splatting for inverse rendering with radiance transfer and reflection. In: ACM Int. Conf. Multimedia. pp. 6888--6897 (2025)

\end{thebibliography}


\setcounter{equation}{0} \renewcommand{\theequation}{S\arabic{equation}}
\setcounter{figure}{0} \renewcommand{\thefigure}{S\arabic{figure}}
\setcounter{table}{0} \renewcommand{\thetable}{S\arabic{table}}

\makeatletter
\renewcommand{\theHequation}{S\theequation}
\renewcommand{\theHfigure}{S\thefigure}
\renewcommand{\theHtable}{S\thetable}
\makeatother

\newcommand{\beginsupplement}{
	\clearpage
	\setcounter{section}{0}
	\renewcommand{\thesection}{\Alph{section}} 
	\renewcommand{\theHsection}{Supp.\arabic{section}} 

	\setcounter{table}{0}
	\renewcommand{\thetable}{S\arabic{table}}%
	\setcounter{figure}{0}
	\renewcommand{\thefigure}{S\arabic{figure}}%
	\setcounter{equation}{0}
	\renewcommand{\theequation}{S\arabic{equation}}

}

\beginsupplement



\section{Additional Methodological Details}
\subsection{Details on Target Pose Reparameterization}
\label{sec:target_pose_detail}
In \cref{sec:vdrefine}, we introduce a 4D reparameterized target-view pose feature as input to the View MLP to facilitate the prediction of view-dependent Gaussian offsets. This section details the formal mathematical derivation of the 3D unit viewing direction and the 1D log-distance from the raw target extrinsic matrix $P^{t \to 1} \in \mathbb{R}^{4 \times 4}$, following the formulation described in the main paper.

\subsubsection{Computing the Camera Center in World Coordinates.}
The predicted target extrinsic matrix $P^{t \to 1}$ represents the world-to-camera (W2C) transformation. For any point $X_w$ in the world (canonical) coordinate system, its position in the camera coordinate system $X_c$ is defined as $X_c = R^{t \to 1}X_w + t^{t \to 1}$, where $R^{t \to 1} \in \mathbb{R}^{3 \times 3}$ is the orthogonal rotation matrix and $t^{t \to 1} \in \mathbb{R}^3$ is the translation vector. Since the camera center $C$ corresponds to the origin $(0, 0, 0)^\top$ in the camera coordinate system, we derive its world-coordinate position $C^{t \to 1}$ as follows:
\begin{equation}
    0 = R^{t \to 1}C^{t \to 1} + t^{t \to 1} \implies C^{t \to 1} = -(R^{t \to 1})^Tt^{t \to 1}.
\end{equation}

\subsubsection{Construction of the 4D Target Pose Feature.}
For each canonical Gaussian with center position $\mu^{v \to 1}_j \in \mathbb{R}^3$ originating from pixel $j$ of context view $v$, we compute the relative pixel-wise displacement vector from the Gaussian center to the target camera center as $\mathbf{d}^v_j = C^{t \to 1} - \mu^{v \to 1}_j$. The View MLP then receives a 4D input $[\mathbf{u}^v_j, l^v_j]$, which consists of a unit viewing direction $\mathbf{u}^v_j \in \mathbb{R}^3$ and a log-scale distance $l^v_j \in \mathbb{R}$, derived from the displacement vector as follows:
\begin{equation}
    \mathbf{u}^v_j = \frac{\mathbf{d}^v_j}{\|\mathbf{d}^v_j\|_2}, \quad l^v_j = \log(\|\mathbf{d}^v_j\|_2 + \epsilon),
    \label{eq:4d_targetpose}
\end{equation}
where $\epsilon$ is a small constant (\eg, $10^{-6}$) for numerical stability.

\subsubsection{Rationale for Log-Scale Transformation.}
The choice of a log-scale for the distance parameter $l$ is motivated by the nonlinear sensitivity inherent in perspective projection. In 3D-to-2D mapping, the visual impact of camera movement is more pronounced for objects at closer ranges. By applying the $\log(\cdot)$ transformation, we compress the dynamic range of distances and maintain higher sensitivity to near-field variations. This allows the View MLP to prioritize parameter refinements for Gaussians where viewpoint changes result in the most significant visual impact, thereby improving the modeling of view-dependent appearance variations.

\subsection{Detailed Formulations of Loss Functions}
\label{sec:loss_detail}
\subsubsection{Image Rendering Loss.} The rendering loss $\mathcal{L}_{render}$ supervises the quality of novel view synthesis by comparing the ground-truth target image $I^t$ with the rendered output $\hat{I}^t$. As described in the main paper, this objective is defined as a weighted combination of an MSE photometric loss and an LPIPS~\cite{zhang2018lpips} perceptual loss:
\begin{equation}
    \mathcal{L}_{render} = \mathcal{L}_{MSE}(I^t, \hat{I}^t) + \lambda_{LPIPS} \mathcal{L}_{LPIPS}(I^t, \hat{I}^t),
    \label{eq:renderingloss}
\end{equation}
where $\lambda_{LPIPS}$ acts as a balancing weight. This formulation encourages the model to achieve both pixel-level photometric accuracy and feature-level perceptual similarity, leading to high-quality rendering results.

\subsubsection{Reprojection Loss.} Since our framework learns 3D Gaussian centers in a canonical space without ground-truth poses, relying solely on the rendering loss can lead to training instability and overfitting to the canonical coordinate frame. To mitigate this and ensure geometric consistency in the absence of ground-truth poses, we utilize $\mathcal{L}_{reproj}$, which provides explicit supervision for both the 3D Gaussian positions and the estimated camera parameters. This loss measures the alignment between the canonical 3D representations and their corresponding 2D observations.

Specifically, for each pixel $p_j^v$ in context view $v$, we project the 3D Gaussian center $\mu_j^{v \to 1}$ from the canonical space to view $v$ using the estimated relative pose $P^{v \to 1}$ and the camera intrinsics $K^v$. The reprojection loss is computed as:
\begin{equation}
    \mathcal{L}_{reproj} = \sum_{v=1}^N  \sum_{j=1}^{H \times W} \|p_j^v - \pi(K^v, P^{v \to 1}, \mu_j^{v \to 1})\|,
    \label{eq:reprojectionloss}
\end{equation}
where $P^{v \to 1} \in \{P_c^{v \to 1}, P_{c+t}^{v \to 1}\}$ and $\pi(\cdot)$ denotes the camera projection function. To maintain consistency, we apply this reprojection loss to the poses estimated from both the context-only branch and the context-with-target branch. This loss stabilizes pose estimation and prevents degenerate solutions where the canonical geometry collapses.

\subsubsection{Total Loss.} The total training objective is the weighted sum of the above terms:
\begin{equation}
    \mathcal{L}_{total} = \mathcal{L}_{render} + \lambda_{reproj}\mathcal{L}_{reproj},
    \label{eq:supp_totalloss}
\end{equation}
where $\lambda_{reproj}$ is a weighting factor. This combined objective enables stable training of the pixel-aligned 3D Gaussians and view-dependent rendering without requiring ground-truth poses. In our experiments, we set the hyperparameters to $\lambda_{LPIPS}=0.05$ and $\lambda_{reproj}=0.001$.

\section{Additional Experiments and Analyses}
\label{sec:additional_exp}
In this section, we provide further ablation studies and additional analyses to thoroughly evaluate our \textbf{ViewSplat} framework. Unless otherwise specified, all experiments in this section are conducted by integrating the ViewSplat framework into the SPFSplat \cite{huang2025spfsplat} baseline.

\subsection{Additional Ablation Studies}
\subsubsection{Ablation on Target Pose Parameterization.}
To empirically validate the effectiveness of our proposed log-scale distance described in \cref{eq:4d_targetpose}, we conduct an ablation study comparing it against a standard linear distance metric $\|\mathbf{d}\|_2$. As summarized in \cref{tab:ablation_targetpose}, the log-scale distance consistently outperforms the linear counterpart in terms of PSNR across both the RE10K~\cite{zhou2018re10k} and ACID~\cite{liu2021acid} datasets.

The improvement in PSNR indicates that the logarithmic transformation provides a more informative signal for the View MLP to refine Gaussian parameters. Specifically, while both metrics share the same normalized viewing direction, the log-scale distance representation exhibits increased sensitivity to near-field variations, allowing the model to better capture complex view-dependent effects that are most prominent at close ranges. These results confirm that reparameterizing distance to align with the nonlinear nature of perspective projection enables more precise modeling of view-dependent effects than a naive linear distance representation.

\begin{table}
    \centering
\caption{Comparison of the proposed log-scale distance against a standard linear distance as input to the View MLP.}
\label{tab:ablation_targetpose}

\setlength{\tabcolsep}{5pt}
\begin{footnotesize}
    \begin{tabular}{l ccc ccc}
        \toprule
        \multirow{2}{*}{\textbf{Method}} & \multicolumn{3}{c}{\textbf{RE10K}} & \multicolumn{3}{c}{\textbf{ACID}} \\
            & PSNR$\uparrow$ & SSIM$\uparrow$ & LPIPS$\downarrow$ & PSNR$\uparrow$ & SSIM$\uparrow$ & LPIPS$\downarrow$ \\ \midrule
        Linear Dist. & 26.296 & 0.857 & 0.144 & 26.618 & 0.790 & 0.177 \\ 
        Log-scale Dist. (\textbf{Ours}) & \textbf{26.317} & 0.857 & 0.144 & \textbf{26.661} & 0.790 & 0.177 \\ \bottomrule
    \end{tabular}
\end{footnotesize}
\end{table}

\subsubsection{Ablation on Reprojection Loss Target.}
We investigate the impact of applying the geometric reprojection loss $\mathcal{L}_{\text{reproj}}$ to the refined centers $\mu + \Delta\mu$ instead of the canonical centers $\mu$. As shown in \cref{tab:ablation_reprojection}, applying the loss to the refined centers slightly degrades performance compared to our default design. This suggests that enforcing rigid geometric supervision on $\mu + \Delta\mu$ restricts the flexibility of the View MLPs. In contrast, supervising the canonical centers keeps $\mu$ as a stable 3D geometric anchor, while allowing $\Delta\mu$ to adaptively model view-dependent appearances without being overly constrained.

\begin{table}
    \centering
\caption{\textbf{Ablation on the reprojection loss target on RE10K.} Supervising canonical centers $\mu$ yields the best performance.}
\label{tab:ablation_reprojection}

\setlength{\tabcolsep}{5pt}
\begin{footnotesize}
    \begin{tabular}{l ccc}
        \toprule
        \textbf{Method} & PSNR$\uparrow$ & SSIM$\uparrow$ & LPIPS$\downarrow$ \\ \midrule
        SPFSplat & 25.484 & 0.847 & 0.153 \\
        +ViewSplat ($\mathcal{L}_{\text{reproj}}$ on $\mu{+}\Delta\mu$) & 26.242 & 0.855 & 0.146 \\
        \textbf{+ViewSplat ($\mathcal{L}_{\text{reproj}}$ on $\mu$) (Ours)} & \textbf{26.317} & \textbf{0.857} & \textbf{0.144} \\ \bottomrule
    \end{tabular}
\end{footnotesize}
\end{table}

\subsubsection{Ablation on Training Strategies.}
\cref{tab:ablation_trainconfig} compares the impact of different initialization and freezing strategies. Our main ViewSplat model, which trains only the view-dependent head while freezing the backbone initialized with SPFSplat weights, achieves superior performance compared to training all parameters initialized from MASt3R weights. For the MASt3R-initialized variant, we adopt learning rates of $1 \times 10^{-5}$ for the encoder and decoder, and $1 \times 10^{-4}$ for the heads. The best performance is achieved by unfreezing the backbone and fine-tuning the entire model (ViewSplat w/ FT), with all learning rates reduced by a factor of 10.

However, jointly training the backbone causes significant additional computational overhead; given that the performance gains are relatively marginal compared to the increased training cost, we adopt the frozen backbone configuration as our default setting. These results indicate that SPFSplat-based initialization provides a robust starting point, while global fine-tuning can further enhance performance at the cost of higher resource requirements.

\begin{table}
    \centering
\caption{Comparison of different training strategies on RE10K. We compare our proposed ViewSplat (SPFSplat-initialized with frozen backbone) against two variants: (1) a fully fine-tuned version starting from our main ViewSplat model, and (2) a model trained from MASt3R pretrained weights without freezing. The \textbf{best} results are highlighted.}
\label{tab:ablation_trainconfig}

\setlength{\tabcolsep}{5pt}
\begin{scriptsize}
    \begin{tabular}{l c c ccc}
        \toprule
        \textbf{Strategy} & \makecell{\textbf{Pretrained} \\ \textbf{Weights}} & \makecell{\textbf{Freeze} \\ \textbf{Pretrained}} & PSNR$\uparrow$ & SSIM$\uparrow$ & LPIPS$\downarrow$ \\ \midrule
        ViewSplat (MASt3R-init) & MASt3R & $\times$ & 26.192 & 0.854 & \textbf{0.140} \\
        ViewSplat & SPFSplat & $\checkmark$ & 26.317 & 0.857 & 0.144 \\
        ViewSplat (w/ FT) & ViewSplat & $\times$ & \textbf{26.459} & \textbf{0.859} & 0.141 \\ \bottomrule
    \end{tabular}
\end{scriptsize}
\end{table}

\subsection{Extended Evaluation and Analysis}
\subsubsection{Quantitative Evaluation of Pose Accuracy and 3D Consistency.}
To verify that ViewSplat achieves high-fidelity rendering through genuine 3D structural refinement rather than 2D overfitting to compensate for inaccurate camera parameters, we evaluate pose accuracy and 3D geometric fidelity.

\textit{Pose Accuracy.} Following previous pose-free methods~\cite{ye2025noposplat,huang2025spfsplat}, we report the area under the cumulative pose error curve (AUC) at thresholds of $5^\circ$, $10^\circ$, and $20^\circ$, where the pose error is defined as the maximum angular error between rotation and translation. As shown in \cref{tab:3d_metric}, ViewSplat has identical pose AUC scores to SPFSplat because the pose head is frozen during training. This confirms that our model does not rely on 2D artifact compensation to bypass inaccurate pose estimation.

\textit{3D Geometric Fidelity.} To assess geometric consistency of the rendered views relative to the ground truth, we measure the first-order depth accuracy ($\delta < 1.25$). Specifically, we use DepthAnything3-Giant~\cite{lin2026da3} to extract dense depth maps from both the rendered and ground-truth target images, and compute the percentage of pixels satisfying $\max(d_{\text{render}}/d_{\text{gt}}, d_{\text{gt}}/d_{\text{render}}) < 1.25$. As shown in \cref{tab:3d_metric}, ViewSplat improves this metric from 0.947 to 0.952, indicating that the proposed view-adaptive updates improve not only image-space fidelity but also depth-level consistency with the target view.

\begin{table}
    \centering
\caption{\textbf{Pose accuracy and 3D consistency on RE10K.} ViewSplat improves rendering and depth consistency while keeping pose accuracy unchanged.}
\label{tab:3d_metric}

\setlength{\tabcolsep}{5pt}
\begin{scriptsize}
    \begin{tabular}{l ccc | ccc | c}
        \toprule
            \multirow{2}{*}{\textbf{Method}} & \multicolumn{3}{c|}{2D Metrics} & \multicolumn{3}{c|}{Pose AUC} & \multicolumn{1}{c}{Depth Consistency} \\ & PSNR$\uparrow$ & SSIM$\uparrow$ & LPIPS$\downarrow$ & $5^\circ{\uparrow}$ & $10^\circ{\uparrow}$ & $20^\circ{\uparrow}$ & $\delta{<}1.25{\uparrow}$\\ \midrule
            SPFSplat & 25.484 & 0.847 & 0.153 & 0.616 & 0.755 & 0.846 & 0.947 \\
            \textbf{+ViewSplat} & \textbf{26.317} & \textbf{0.857} & \textbf{0.144} & 0.616 & 0.755 & 0.846 & \textbf{0.952} \\ \bottomrule
    \end{tabular}
\end{scriptsize}
\end{table}

\subsubsection{Context View Scaling}
We evaluate the scalability of ViewSplat by increasing the number of input context views to denser settings ($N=5$ and $10$), as summarized in \cref{tab:multi_view}. The results show that ViewSplat consistently improves reconstruction quality over the SPFSplat baseline as the number of context views increases.

This scaling behavior suggests that denser context views provide a stronger global geometric foundation, allowing our scene-conditioned hypernetwork to predict more effective view-dependent residual adjustments. These results validate that ViewSplat scales favorably from sparse context pairs to denser multi-view inputs while delivering increasingly large gains as context coverage improves.

\subsubsection{Generalization of ViewSplat on an Alternative Backbone.}
To examine whether ViewSplat is a general architectural paradigm rather than a backbone-specific modification, we further evaluate its plug-and-play applicability on an independent contemporary pose-free architecture, YoNoSplat~\cite{ye2026yonosplat}. As summarized in \cref{tab:yono_viewsplat}, incorporating our view-adaptive splatting module into YoNoSplat consistently improves performance under both 2-view and 6-view settings. These results demonstrate that our module can refine initial estimates and enhance view-dependent details across distinct backbone architectures, supporting the generality and transferability of the proposed view-adaptive design.

\begin{table}
    \centering
\begin{minipage}{0.42\textwidth}
    \centering
    \caption{Quantitative evaluation of context view scaling on RE10K.}
    \label{tab:multi_view}
    \setlength{\tabcolsep}{4pt}
    \resizebox{\linewidth}{!}{%
        \begin{tabular}{l ccc}
            \toprule
            \textbf{Method} & PSNR $\uparrow$ & SSIM $\uparrow$ & LPIPS $\downarrow$ \\ \midrule
            \textit{5 views} & & & \\
            SPFSplat & 26.891 & 0.875 & \textbf{0.122} \\
            \textbf{+ViewSplat} & \textbf{27.607} & \textbf{0.885} & 0.124 \\ \midrule
            \textit{10 views} & & & \\
            SPFSplat & 27.159 & 0.880 & 0.115 \\
            \textbf{+ViewSplat} & \textbf{29.010} & \textbf{0.905} & \textbf{0.098} \\ \bottomrule
    \end{tabular}
    }
\end{minipage}%
\hfill%
\begin{minipage}{0.56\textwidth}
    \centering
    \caption{Backbone-agnostic generalization of ViewSplat on RE10K.}
    \label{tab:yono_viewsplat}
    \setlength{\tabcolsep}{4pt}
    \resizebox{\linewidth}{!}{
        \begin{tabular}{l ccc}
            \toprule
            \textbf{Method} & PSNR $\uparrow$ & SSIM $\uparrow$ & LPIPS $\downarrow$ \\ \midrule
            YoNoSplat (2 views) & 22.847 & 0.757 & 0.181 \\
            \textbf{+ViewSplat (2 views)} & \textbf{23.354} & \textbf{0.766} & \textbf{0.171} \\
            YoNoSplat (6 views) & 25.071 & 0.814 & 0.114 \\
            \textbf{+ViewSplat (6 views)} & \textbf{25.975} & \textbf{0.830} & \textbf{0.102}  \\ \bottomrule
        \end{tabular}
    }
\end{minipage}
\end{table}

\subsubsection{Spatial Breakdown of Visual Refinements.}
\begin{figure}
    \centering
    \includegraphics[width=0.9\linewidth]{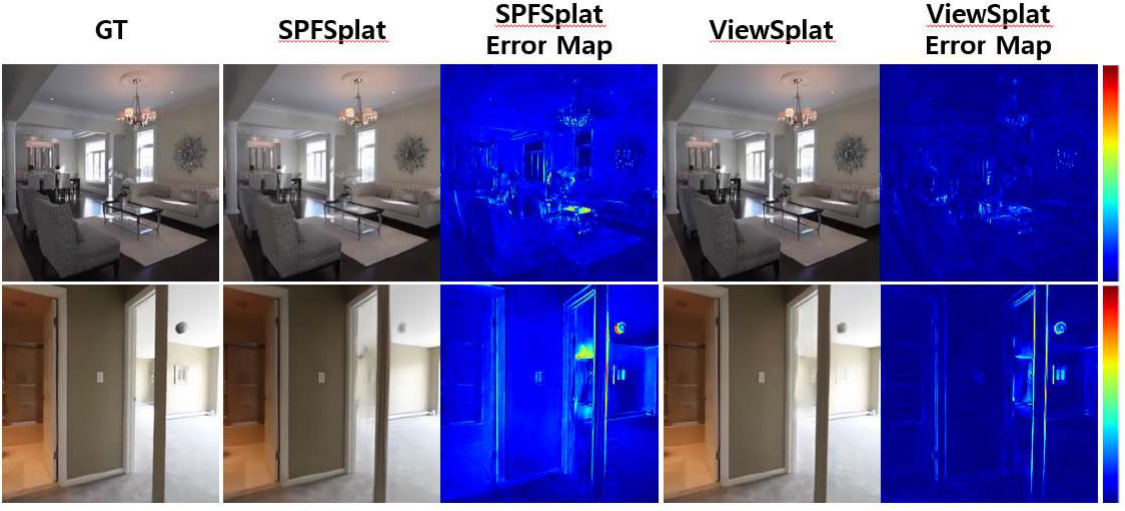}
    \caption{\textbf{Spatial breakdown of rendering errors.} Pixel-wise error maps (red: high, blue: low) show that ViewSplat mainly reduces errors around specular regions and high-frequency boundaries while keeping diffuse regions stable.}
    \label{fig:error_map}
\end{figure}

We further analyze ViewSplat using pixel-wise error maps in \cref{fig:error_map}. Compared to the static SPFSplat baseline, ViewSplat reduces errors mainly around challenging regions, such as specular reflections and high-frequency geometric boundaries, while leaving diffuse regions largely stable. This spatially selective behavior suggests that the scene-conditioned View MLPs perform targeted pose-conditioned refinements where static Gaussian primitives are insufficient.

\subsubsection{Detailed Results on Varying Camera Overlap.}
\label{sec:camera_overlap_detail}
We present the detailed quantitative results on the RE10K and ACID datasets under varying levels of camera overlap. Following prior work~\cite{ye2025noposplat,huang2025spfsplat}, we categorize input pairs into three groups based on image overlap ratios: small (0.05--0.3), medium (0.3--0.55), and large (0.55--0.8), which are determined using a pretrained dense image matching method \cite{edstedt2024roma}.

The detailed results are presented in \cref{tab:supp_nvs_re10k} and \cref{tab:supp_nvs_acid}. Consistent with the average performance, our \textbf{ViewSplat} framework yields robust and consistent gains for all SPFSplat variants across all overlap ratios.

\begin{table}
    \centering
\caption{Detailed performance of novel view synthesis on the RE10K dataset under varying camera overlap ratios.}
\label{tab:supp_nvs_re10k}

\setlength{\tabcolsep}{2.5pt}
\resizebox{\linewidth}{!}{
    \begin{tabular}{l ccc ccc ccc ccc}
        \toprule
        \multirow{2}{*}{\textbf{Method}} & \multicolumn{3}{c}{\textbf{Small}} & \multicolumn{3}{c}{\textbf{Medium}} & \multicolumn{3}{c}{\textbf{Large}} & \multicolumn{3}{c}{\textbf{Average}} \\
            & PSNR$\uparrow$ & SSIM$\uparrow$ & LPIPS$\downarrow$ & PSNR$\uparrow$ & SSIM$\uparrow$ & LPIPS$\downarrow$ & PSNR$\uparrow$ & SSIM$\uparrow$ & LPIPS$\downarrow$ & PSNR$\uparrow$ & SSIM$\uparrow$ & LPIPS$\downarrow$ \\ \midrule
        \textit{Pose-Required} & & & & & & & & & & & & \\
            pixelSplat & 20.277 & 0.719 & 0.265 & 23.726 & 0.811 & 0.180 & 27.152 & 0.880 & 0.121 & 23.859 & 0.808 & 0.184 \\
            MVSplat & 20.371 & 0.725 & 0.250 & 23.808 & 0.814 & 0.172 & 27.466 & 0.885 & 0.115 & 24.012 & 0.812 & 0.175 \\ \midrule
        \textit{Supervised Pose-Free} & & & & & & & & & & & & \\
            CoPoNeRF & 17.393 & 0.585 & 0.462 & 18.813 & 0.616 & 0.392 & 20.464 & 0.652 & 0.318 & 18.938 & 0.619 & 0.388 \\
            Splatt3R & 17.789 & 0.582 & 0.375 & 18.828 & 0.607 & 0.330 & 19.243 & 0.593 & 0.317 & 18.688 & 0.337 & 0.596 \\
            NoPoSplat* & 22.514 & 0.784 & 0.210 & 24.899 & 0.839 & 0.160 & 27.411 & 0.883 & 0.119 & 25.033 & 0.838 & 0.160 \\ \midrule
        \textit{Self-Supervised Pose-Free} & & & & & & & & & & & & \\
            SelfSplat & 14.828 & 0.543 & 0.469 & 18.857 & 0.679 & 0.328 & 23.338 & 0.798 & 0.208 & 19.152 & 0.680 & 0.328 \\
            PF3plat & 18.358 & 0.668 & 0.298 & 20.953 & 0.741 & 0.231 & 23.491 & 0.795 & 0.179 & 21.042 & 0.739 & 0.233 \\ \midrule
            SPFSplat & 22.897 & 0.792 & 0.201 & 25.334 & 0.847 & 0.153 & 27.947 & 0.894 & 0.110 & 25.484 & 0.847 & 0.153 \\
            \textbf{+ViewSplat} & \textbf{23.562} & \textbf{0.802} & \textbf{0.191} & \textbf{26.193} & \textbf{0.858} & \textbf{0.143} & \textbf{28.886} & \textbf{0.904} & \textbf{0.104} & \textbf{26.317} & \textbf{0.857} & \textbf{0.144} \\ \midrule
            SPFSplatV2 & 23.123 & 0.800 & 0.195 & 25.542 & 0.853 & 0.149 & 28.143 & 0.897 & 0.110 & 25.693 & 0.853 & 0.149 \\
            \textbf{+ViewSplat} & \textbf{23.761} & \textbf{0.808} & \textbf{0.185} & \textbf{26.353} & \textbf{0.862} & \textbf{0.139} & \textbf{28.978} & \textbf{0.906} & \textbf{0.102} & \textbf{26.468} & \textbf{0.861} & \textbf{0.140} \\ \midrule
            SPFSplatV2-L & 23.138 & 0.804 & 0.184 & 25.518 & 0.856 & 0.136 & 28.081 & 0.899 & 0.099 & 25.668 & 0.855 & 0.137 \\
            \textbf{+ViewSplat}  & \textbf{24.129} & \textbf{0.819} & \textbf{0.168} & \textbf{26.686} & \textbf{0.871} & \textbf{0.123} & \textbf{29.274} & \textbf{0.912} & \textbf{0.088} & \textbf{26.798} & \textbf{0.870} & \textbf{0.124} \\ \bottomrule
    \end{tabular}
    }
\end{table}

\begin{table}
    \centering
\caption{Detailed performance of novel view synthesis on the ACID dataset under varying camera overlap ratios.}
\label{tab:supp_nvs_acid}

\setlength{\tabcolsep}{2.5pt}
\resizebox{\linewidth}{!}{
    \begin{tabular}{l ccc ccc ccc ccc}
        \toprule
        \multirow{2}{*}{\textbf{Method}} & \multicolumn{3}{c}{\textbf{Small}} & \multicolumn{3}{c}{\textbf{Medium}} & \multicolumn{3}{c}{\textbf{Large}} & \multicolumn{3}{c}{\textbf{Average}} \\
            & PSNR$\uparrow$ & SSIM$\uparrow$ & LPIPS$\downarrow$ & PSNR$\uparrow$ & SSIM$\uparrow$ & LPIPS$\downarrow$ & PSNR$\uparrow$ & SSIM$\uparrow$ & LPIPS$\downarrow$ & PSNR$\uparrow$ & SSIM$\uparrow$ & LPIPS$\downarrow$ \\ \midrule
        \textit{Pose-Required} & & & & & & & & & & & & \\
            pixelSplat & 22.088 & 0.655 & 0.284 & 25.525 & 0.777 & 0.197 & 28.527 & 0.854 & 0.139 & 25.889 & 0.780 & 0.194 \\
            MVSplat & 21.412 & 0.640 & 0.290 & 25.150 & 0.772 & 0.198 & 28.457 & 0.854 & 0.137 & 25.561 & 0.775 & 0.195 \\ \midrule
        \textit{Supervised Pose-Free} & & & & & & & & & & & & \\
            CoPoNeRF & 18.651 & 0.551 & 0.485 & 20.654 & 0.595 & 0.418 & 22.654 & 0.652 & 0.343 & 20.950 & 0.606 & 0.406 \\
            Splatt3R & 17.419 & 0.501 & 0.434 & 18.257 & 0.514 & 0.405 & 18.134 & 0.508 & 0.395 & 18.060 & 0.510 & 0.407 \\
            NoPoSplat* & 23.087 & 0.685 & 0.258 & 25.624 & 0.777 & 0.193 & 28.043 & 0.841 & 0.144 & 25.961 & 0.781 & 0.189 \\ \midrule
        \textit{Self-Supervised Pose-Free} & & & & & & & & & & & & \\
            SelfSplat & 18.301 & 0.568 & 0.408 & 21.375 & 0.676 & 0.314 & 25.219 & 0.792 & 0.214 & 22.089 & 0.694 & 0.298 \\
            PF3plat & 18.112 & 0.537 & 0.376 & 20.732 & 0.615 & 0.307 & 23.607 & 0.710 & 0.228 & 21.206 & 0.632 & 0.293 \\ \midrule
            SPFSplat & 22.667 & 0.665 & 0.262 & 25.620 & 0.773 & 0.192 & 28.607 & 0.856 & 0.136 & 26.070 & 0.781 & 0.186 \\
            \textbf{+ViewSplat} & \textbf{23.035} & \textbf{0.671} & \textbf{0.255} & \textbf{26.161} & \textbf{0.782} & \textbf{0.184} & \textbf{29.395} & \textbf{0.869} & \textbf{0.124} & \textbf{26.661} & \textbf{0.790} & \textbf{0.177} \\ \midrule
            SPFSplatV2 & 22.944 & 0.679 & 0.255 & 25.849 & 0.784 & 0.187 & 28.766 & 0.862 & 0.133 & 26.284 & 0.791 & 0.182 \\
            \textbf{+ViewSplat} & \textbf{23.264} & \textbf{0.685} & \textbf{0.249} & \textbf{26.398} & \textbf{0.794} & \textbf{0.179} & \textbf{29.561} & \textbf{0.875} & \textbf{0.122} & \textbf{26.873} & \textbf{0.801} & \textbf{0.173} \\ \midrule
            SPFSplatV2-L & 23.640 & 0.706 & 0.225 & 26.272 & 0.801 & 0.166 & 28.938 & 0.868 & 0.120 & 26.674 & 0.806 & 0.162 \\
            \textbf{+ViewSplat} & \textbf{24.234} & \textbf{0.718} & \textbf{0.214} & \textbf{27.090} & \textbf{0.815} & \textbf{0.154} & \textbf{29.931} & \textbf{0.885} & \textbf{0.106} & \textbf{27.509} & \textbf{0.820} & \textbf{0.149} \\ \bottomrule
    \end{tabular}
    }
\end{table}

\clearpage

\section{More Qualitative Results}
\label{sec:more_qualitative}
\begin{figure}
    \centering
    \includegraphics[width=1\linewidth]{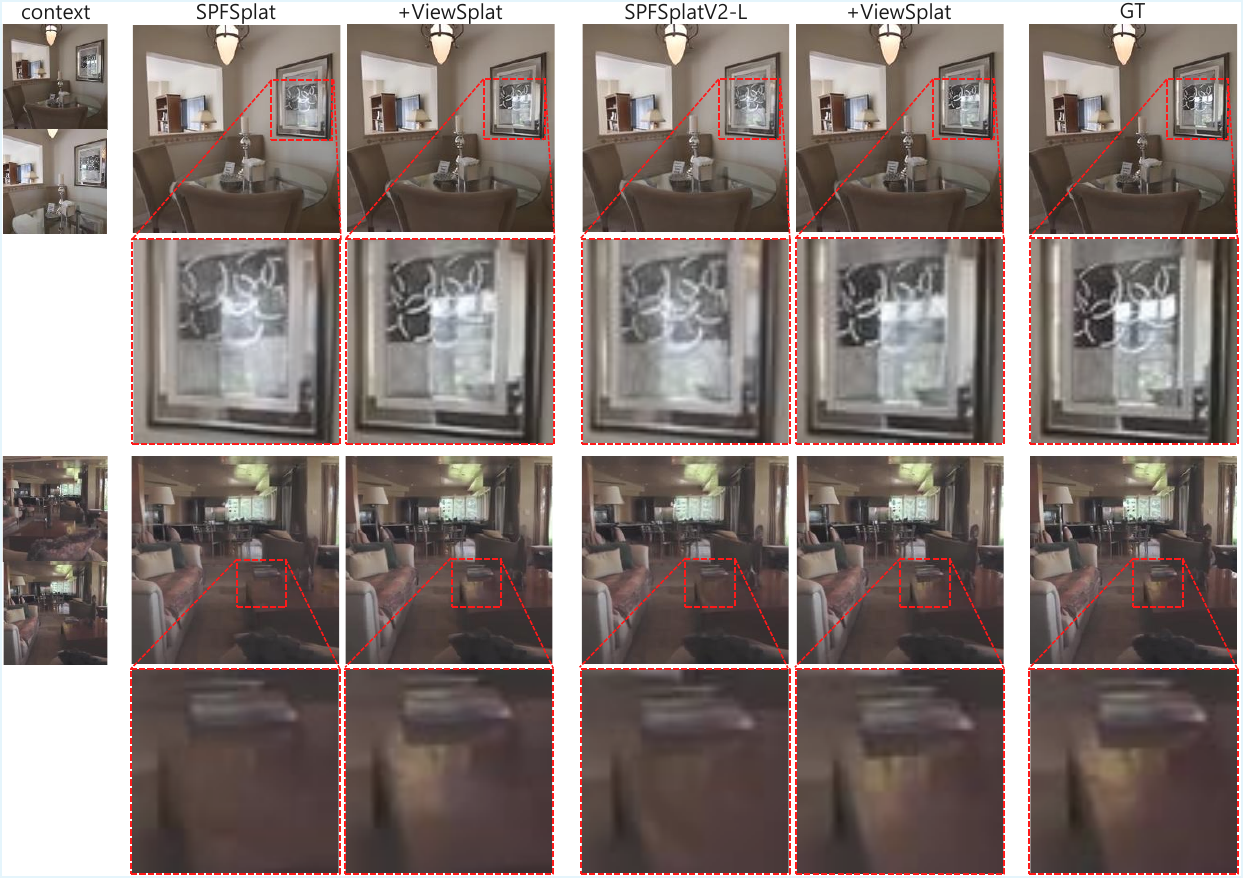}
    \caption{More qualitative results on RE10K with large image overlap.}
    \label{fig:qualitative_re10k_large}
\end{figure}

\begin{figure}
    \centering
    \includegraphics[width=1\linewidth]{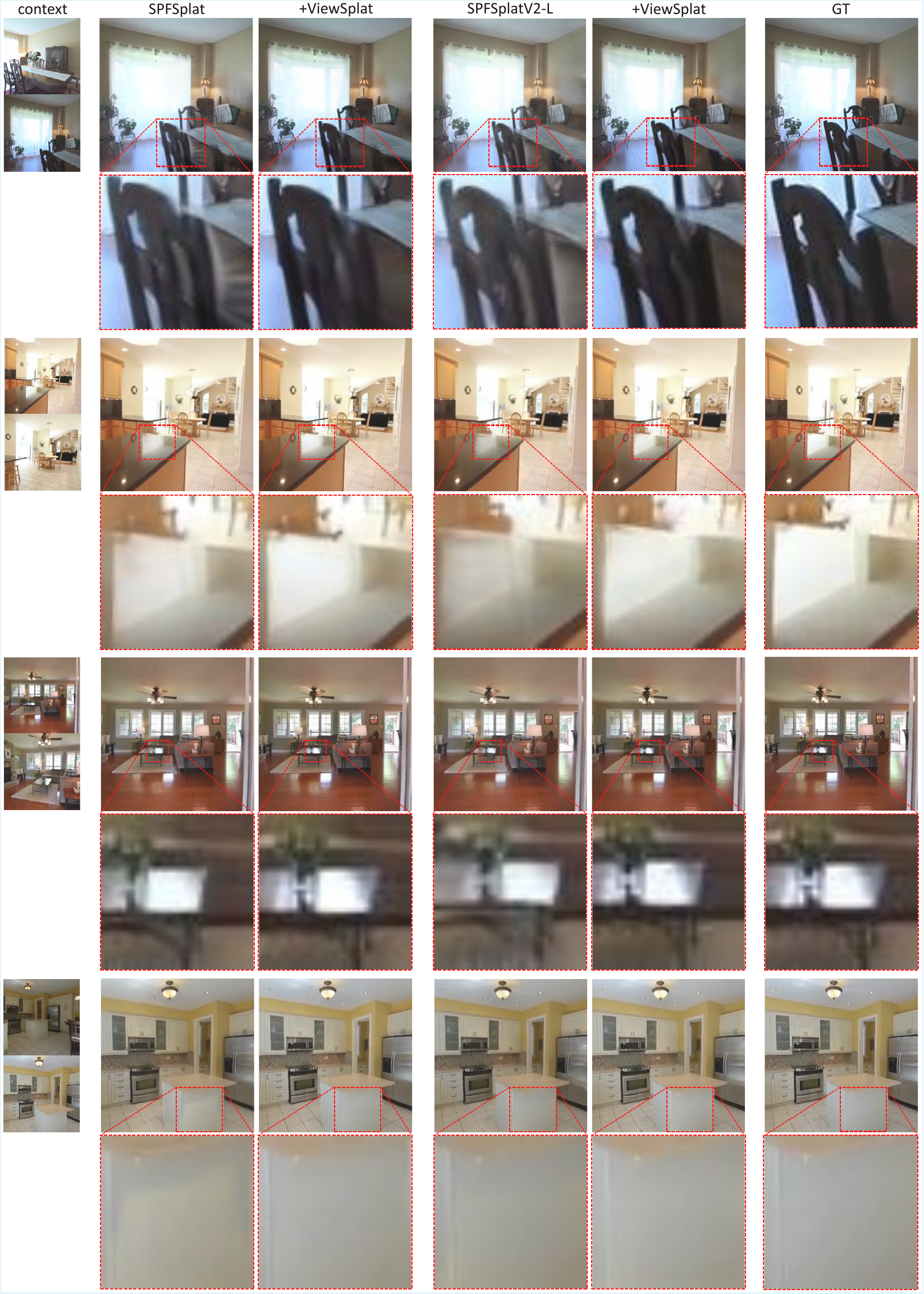}
    \caption{More qualitative results on RE10K with medium image overlap.}
    \label{fig:qualitative_re10k_medium1}
\end{figure}

\begin{figure}
    \centering
    \includegraphics[width=1\linewidth]{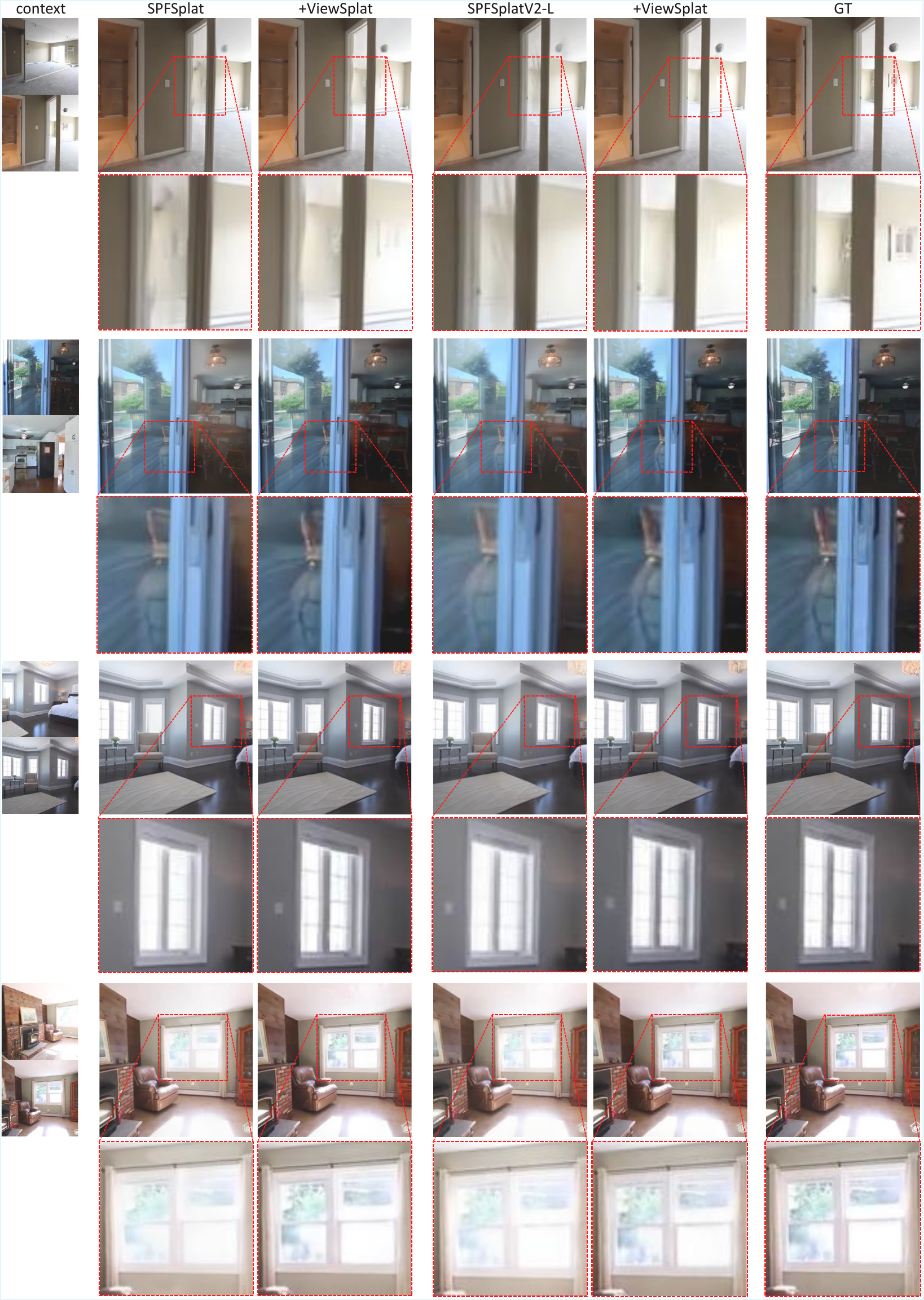}
    \caption{More qualitative results on RE10K with medium image overlap.}
    \label{fig:qualitative_re10k_medium2}
\end{figure}

\begin{figure}
    \centering
    \includegraphics[width=1\linewidth]{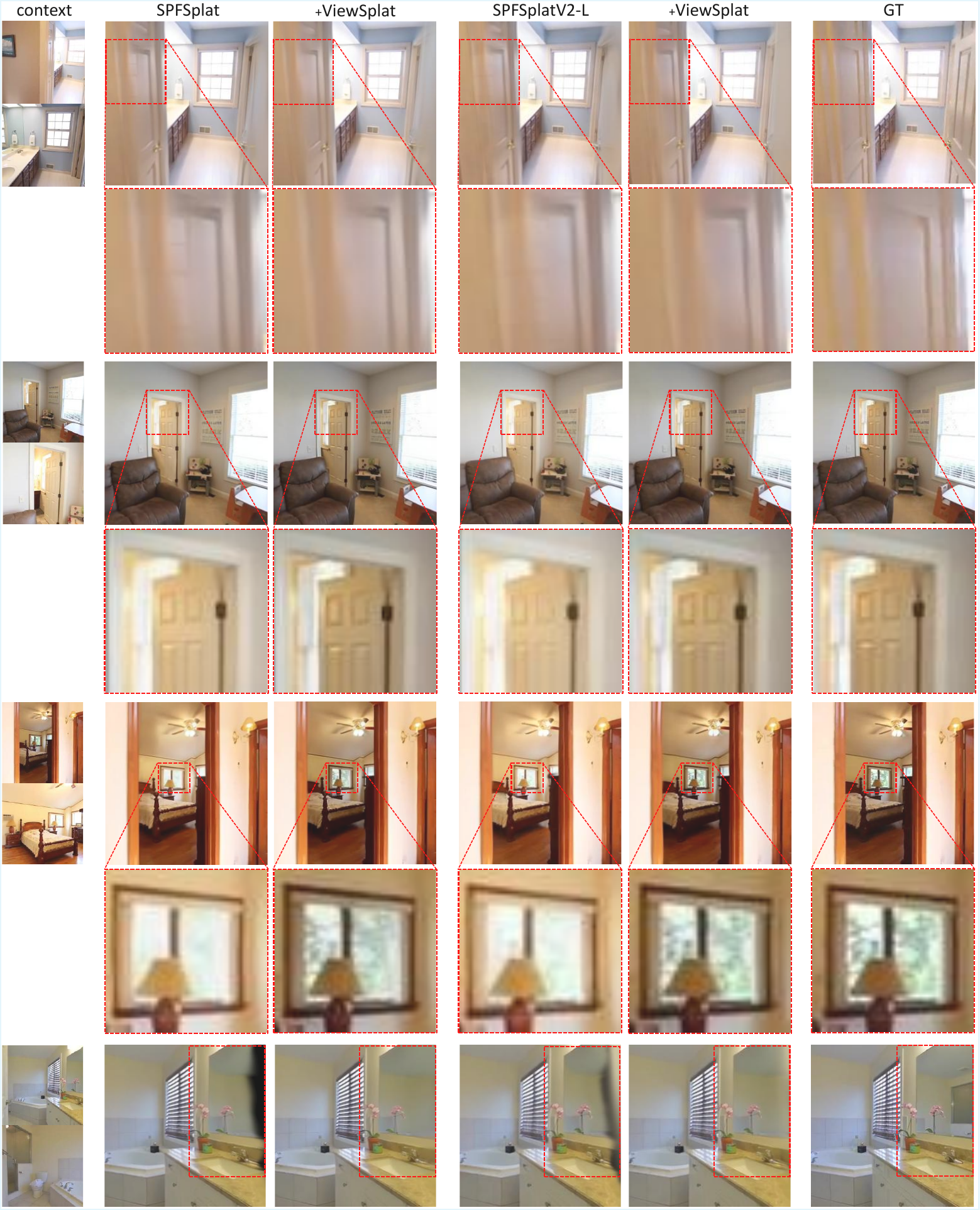}
    \caption{More qualitative results on RE10K with small image overlap.}
    \label{fig:qualitative_re10k_small}
\end{figure}

\end{document}